\documentclass{article}

\usepackage{graphicx}
\usepackage{amsmath}
\usepackage{amsthm}
\usepackage{natbib}
\usepackage{amsfonts}
\usepackage{hyperref}
\usepackage{authblk}
\usepackage{geometry}
\usepackage{float}
\usepackage{bm}

\usepackage{listings}
\usepackage{xcolor}
\definecolor{codekeyword}{HTML}{8B008B}
\definecolor{codestring}{HTML}{CD5555}
\definecolor{codecomment}{HTML}{228B22}
\definecolor{codeidentifier}{HTML}{00688B}
\lstdefinelanguage{Rlang}{
  keywords={if,else,for,while,repeat,function,return,break,next,in,TRUE,FALSE,NULL,NA,Inf,NaN},
  sensitive=true,
  morecomment=[l]{\#},
  morestring=[b]",
  morestring=[b]'
}
\usepackage{mdframed}

\newgeometry{
    textheight=9in,
    textwidth=5.5in,
    top=1in,
    headheight=12pt,
    headsep=25pt,
    footskip=30pt
}

\title{Learning density ratios in causal inference using Bregman--Riesz regression}
\date{\today}
\author[1]{Oliver J.~Hines}
\author[1]{Caleb H.~Miles}
\affil[1]{Columbia University, New York, NY, USA}

\newcommand{\dd}{\mathrm{d}}
\newcommand{\E}{\mathbb{E}}
\newcommand{\h}{\mathbb{H}}

\newcommand{\Var}{\mathrm{var}}
\newcommand{\n}{\mathcal{N}}

\DeclareMathOperator*{\argmin}{arg\,min}

\DeclareMathOperator{\Tr}{Tr}

\theoremstyle{plain}
\newtheorem{theorem}{Theorem}%[section]
\newtheorem{corollary}{Corollary}[theorem]
\newtheorem{lemma}{Lemma}

\theoremstyle{definition}

\newtheorem{example}{Example}
\newtheorem{setting}{Setting}
\newtheorem{augmentation_example}{Augmentation Example}
\newtheorem{divergence}{Divergence}
\newmdtheoremenv[innertopmargin=0.5em,innerbottommargin=0.5em,innerleftmargin=0.8em,innerrightmargin=0.8em]{algorithm}{Algorithm}

\theoremstyle{remark}
\newtheorem{remark}{Remark}

\begin{document}

\maketitle

\begin{abstract}
    The ratio of two probability density functions is a fundamental quantity that appears in many areas of statistics and machine learning, including causal inference, reinforcement learning, covariate shift, outlier detection, independence testing, importance sampling, and diffusion modeling. Naively estimating the numerator and denominator densities separately using, e.g., kernel density estimators, can lead to unstable performance and suffer from the curse of dimensionality as the number of covariates increases. For this reason, several methods have been developed for estimating the density ratio directly based on (a) Bregman divergences or (b) recasting the density ratio as the odds in a probabilistic classification model that predicts whether an observation is sampled from the numerator or denominator distribution. Additionally, the density ratio can be viewed as the Riesz representer of a continuous linear map, making it amenable to estimation via (c) minimization of the so-called Riesz loss, which was developed to learn the Riesz representer in the Riesz regression procedure in causal inference. In this paper we show that all three of these methods can be unified in a common framework, which we call Bregman--Riesz regression. We further show how data augmentation techniques can be used to apply density ratio learning methods to causal problems, where the numerator distribution typically represents an unobserved intervention.
    We show through simulations how the choice of Bregman divergence and data augmentation strategy can affect the performance of the resulting density ratio learner.
    A Python package is provided for researchers to apply Bregman--Riesz regression in practice using gradient boosting, neural networks, and kernel methods.
\end{abstract}

\section{Introduction}

The ratio of two probability density functions (density ratio) is a fundamental quantity that appears in many areas of statistics and machine learning.
In this article we focus on applications in causal inference; however, density ratios also appear when:
reweighting data due to sample selection biases \citep{huang_correcting_2007};
detecting outliers / covariate shift in test data \citep{hido_statistical_2011};
detecting change points in time series \citep{liu_change-point_2013};
estimating mutual information \citep{nguyen_estimating_2007,belghazi_mutual_2018};
and in generative diffusion modeling \citep{meng_concrete_2023,lou_discrete_2024}.
One reason for the ubiquity of density ratios is that they can be viewed as non-negative importance weights when expressing expectations with respect to the numerator distribution $P_1$, as weighted expectations with respect to the denominator distribution $P_0$.
For this reason, density ratios appear frequently in causal inference, where $P_1$ represents a target intervention distribution and $P_0$ represents the observed data generating distribution. For instance, in policy learning / offline reinforcement learning, one is interested in computing the expected loss if a treatment/action had been assigned according to a counterfactual decision rule which may depend on the natural value of the treatment \citep{diaz_population_2012,haneuse_estimation_2013,kallus_double_2020,diaz_nonparametric_2023} or its distribution \citep{kennedy_nonparametric_2019}.

Naively learning the numerator and denominator densities separately then taking their ratio can lead to unstable estimates of the density ratio. A well-known example of this phenomenon in causal inference occurs when using inverse probability weighting (IPW) strategies to estimate the mean $\E_{P_0}[Y^1]$ of an outcome $Y^a$ that would be observed under an intervention that assigns a binary treatment $A=a$ uniformly in a population.
Under standard causal assumptions of consistency, positivity, and conditional exchangeability given predictors $W$, the target estimand is identified by $\E_{P_0}[Y \alpha_0(X)]$ where, for $X=(A,W)$, the quantity $\alpha_0(x) = p_1(x) / p_0(x)$ is an importance weight that represents the ratio between the density of the observed data-generating distribution $p_0(x) = p_{A,W}(a, w)$ and the density of the intervention distribution $p_1(x) = \mathbb{I}(a = 1)p_W(w)$. (We use the terms \textit{density} and \textit{probability mass function} interchangeably throughout.)
This identification result suggests the IPW estimator
\begin{align*}
n^{-1} \sum_{i=1}^n \underbrace{\frac{\mathbb{I}(a_i = 1)}{\hat{p}(w_i)}}_{\text{importance weight}} y_i
\end{align*}
where $\hat{p}(w_i)$ is an estimator of the propensity score $p_{A\mid W}(1\mid w_i)$. However, the IPW estimator is known to be overly sensitive to small errors in the estimated propensity score $\hat{p}(w_i)$ when $a_i = 1$ and $p_{A\mid W}(1\mid w_i)$ is close to zero, since the importance weight above can take extreme values in this case. Motivated by this and similar issues for continuous densities, several methods have been proposed to estimate density ratios directly from data, without having to learn individual component models such as the propensity score model above.

Rather than focusing on density ratios specifically, developments in the causal inference literature have instead focused on learning so-called \textit{Riesz representers}, which are signed weighting functions corresponding to certain linear statistical estimands, with the density ratio being an example of a Riesz representer.
A growing body of such literature approximates the Riesz representer as a \textit{balancing weight}, which minimizes some measure of the worst-case imbalance over a set of test functions, while penalizing the complexity of the weights. We refer the reader to \cite{ben-michael_balancing_2021} for a review, with early proposals including:
generalized regression estimators \citep{deville_calibration_1992};
kernel mean matching \citep{huang_correcting_2007};
entropy balancing \citep{hainmueller_entropy_2012};
covariate balance propensity scores \citep{imai_covariate_2014,fong_covariate_2018};
stable balancing weights \citep{Zubizarreta2015};
and more recent proposals combining balancing weight estimation with outcome regression estimation, such as in:
approximate residual balancing \citep{athey_approximate_2018};
augmented minimax linear estimation \citep{Hirshberg2021,Dikkala2020};
and recent work on duality by \cite{bruns-smith_outcome_2022}. % and \cite{bruns-smith_augmented_2024}.

Balancing weight methods successfully alleviate the problem of extreme weights in importance weighted estimators and circumvent the need to model components of the Riesz representer / density ratio. However, they also have several drawbacks: (a) when the test function set is not sufficiently rich there is no guarantee that the resulting weights provide a good approximation to the target Riesz representer; (b) when the test function set is a rich function set, the resulting minimax problem (minimize over the weights, maximize over the test functions) becomes computationally challenging (see, for example, the adversarial learner of \cite{chernozhukov_adversarial_2024}); and (c) several balancing weight approaches estimate only the weights evaluated at observations $\hat{\alpha}_i \approx \alpha_0(x_i)$ rather than learn the function $\alpha_0$ itself. Such methods are not amenable to making out-of-sample predictions, meaning that model selection/validation techniques, such as cross-validation, are not possible.

To overcome these issues, we focus on methods which learn Riesz representers via empirical risk minimization, building on previous work by \cite{chernozhukov_automatic_2024} who propose \textit{Riesz regression} as a general Riesz representer learning approach, where an empirical risk is constructed based on the mean squared error loss. A special case of Riesz regression can also be found in earlier work on score matching by \cite{hyvarinen_estimation_2005}.

Although Riesz regression has become a popular paradigm in causal inference using machine learning, several other density ratio learning methodologies have been proposed when one observes $n_1$ samples from the numerator distribution $P_1$ and $n_0$ samples from the denominator distribution $P_0$. In particular, empirical risk minimization approaches were pioneered through the Bregman divergence framework of \cite{gutmann_bregman_2011} and \cite{sugiyama_density-ratio_2012}, with Bregman divergences being a fundamental class of divergences, due to \cite{bregman_relaxation_1967}, that generalize the squared Euclidean distance. Their framework includes as special cases the Kullback--Leibler Importance Estimation Procedure (KLIEP) \citep{sugiyama_direct_2007}, which minimizes the Kullback--Leibler divergence, and unconstrained Least Squares Importance Fitting (uLSIF) \citep{kanamori_least-squares_2009}, which minimizes a squared error loss using the same decomposition as in Riesz regression.
Another common density ratio learning approach is to recast the density ratio as the odds from a probabilistic classifier that learns whether an observation is drawn from $P_1$ or $P_0$ \citep{qin_inferences_1998,cheng_semiparametric_2004,bickel_multi-task_2008}. Perhaps surprisingly, however, probabilistic classification via empirical risk minimization is equivalent to learning the density ratio via an empirical Bregman divergence for a broad class of `proper' classification losses that includes the binary cross-entropy and exponential losses \citep{menon_linking_2016}.

One reason we believe that general density ratio learning methods have not been more widely adopted in causal machine learning is due to the requirement to observe samples from $P_1$, which in causal settings usually represents an intervention distribution that is not directly sampled from. Generating such samples requires the use of data augmentation techniques, along with careful consideration of how the intervention distribution is constructed. However, examples of such data augmentation strategies being used in causal inference do exist. For example, \cite{diaz_nonparametric_2023} generate augmented data for longitudinal modified treatment policies and learn their density ratio of interest via probabilistic classification. \cite{Lee2025} generate augmented data for policy / shift interventions, and learn their density ratio via Riesz regression. The latter work is motivated by a separate concern of Riesz regression: without data augmentation, Riesz regression is usually not amenable for use in gradient boosting algorithms because the unit loss can depend on the Riesz representer value at more than one point in the covariate space. For example, when learning the importance weights for the IPW estimator above, the loss for the $i$th observation depends on the values $\alpha(a_i, w_i)$ and $\alpha(1, w_i)$, for a candidate model $\alpha$.

Our main contribution is to propose Bregman--Riesz regression as an extension to Riesz regression, which includes several density ratio learning methods in a unifying framework based on Bregman divergences.
We show why data augmentation methods are often required when applying density ratio learning methods to causal problems, where data from the intervention distribution is not available. The need to develop estimand-specific data augmentation is arguably a barrier to the adoption of density ratio learning methods in causal inference. We address this by demonstrating several augmentation strategies for common counterfactual interventions (modified treatment policies, stabilized weights, natural mediation interventions). We also provide practical recommendations based on a numerical study of the effect of different losses and augmentation strategies when learning causal density ratios.
Our experiments use neural network models, gradient-boosted tree models, and kernel methods, with our implementations made available via a Python package at \url{https://github.com/CI-NYC/densityratios}.

In Section \ref{sect:setup} we describe the Riesz representer setup and several canonical causal examples.
In Section \ref{sect:breg} we introduce the Bregman--Riesz regression, focusing on four divergences: the least-squares, Kullback--Leibler, negative binomial, and Itakura--Saito.
Section \ref{sect:dr} shows how Bregman--Riesz regression specializes to density ratio learning, which requires data augmentation strategies for the causal examples we consider.
These are discussed in Section \ref{sect:augment}.
In Section \ref{sect:prob} we establish a formal connection between Bregman--Riesz regression and probabilistic classification, enabling implementation via standard outcome regression.
Numerical experiments are presented in Section \ref{sect:simulations}, related work is reviewed in Section \ref{sect:related}, and Section \ref{sect:discussion} concludes.

\section{Methods}
\label{sect:Methods}

\subsection{Problem description}
\label{sect:setup}

Consider a random vector $X \in \mathcal{X}$ distributed according to an unknown distribution $P_0$. Let $\mathcal{H}$ be a Hilbert space of real-valued functions $f:\mathcal{X} \to \mathbb{R}$ with inner product $\langle f, g\rangle \equiv \E_{P_0}[f(X) g(X)]$ for $f, g \in \mathcal{H}$ and norm $\|f\|_{\mathcal{H}} \equiv \langle f, f \rangle^{1/2}$. Letting $\h: \mathcal{H} \to \mathbb{R}$ be a bounded linear functional, the Riesz representation theorem implies that there exists a unique function $\alpha_0 \in \mathcal{H}$, called the Riesz representation of $\mathcal{H}$, or Riesz representer for short, with the property that $\h(f) = \langle f, \alpha_0 \rangle$ for all $f\in \mathcal{H}$. The following are two canonical examples of this setup in the statistical literature.

\begin{setting}[Density Ratio]
\label{setting:dr}
Let $P_1$ be an alternative distribution over $\mathcal{X}$ that is absolutely continuous with respect to $P_0$, i.e., the support of $P_1$ is contained in the support of $P_0$. The map $\h(f) = \E_{P_1}[f(X)]$ has a Riesz representer that is the density ratio $\alpha_0(x) = \dd P_1(x) / \dd P_0(x)$, also known as the Radon--Nikodym derivative in probability theory. The density ratio is more commonly written as $\alpha_0(x) = p_1(x) / p_0(x)$ where $p_1(x) = \dd P_1(x) / \dd \nu(x)$ and $p_0(x) = \dd P_0(x) / \dd \nu(x)$ are densities with respect to a dominating measure $\nu$, e.g., the Lebesgue measure or the counting measure. In this way, the density ratio is agnostic as to whether $\mathcal{X}$ is discrete or continuous.
\end{setting}

\begin{setting}[Average Linear Functional]
\label{setting:alf}
Let $\h(f) = \E_{P_0}[m(f, X, Y)]$, where $m: \mathcal{H} \times\mathcal{X} \times \mathbb{R} \to \mathbb{R}$ is a known functional that is linear in its first argument, and $Y \in \mathbb{R}$ is an outcome with $(Y, X) \sim P_0$.
The Riesz representer is the unique function $\alpha_0 \in \mathcal{H}$ such that $\E_{P_0}[m(f, X, Y)] = \langle f, \alpha_0\rangle$.
This setting has been studied by \cite{Hirshberg2021}, \cite{chernozhukov_locally_2022}, and \cite{chernozhukov_automatic_2024}, with primary interest in the debiased estimation of $\h(\mu_0)$, where $\mu_0(x) = \E_{P_0}[Y  \mid X=x]$ and it is assumed that $\mu_0 \in \mathcal{H}$.
\end{setting}

One key difference between these two settings is that in Setting \ref{setting:dr}, the density ratio $\alpha_0(x) \geq 0$ is a nonnegative importance weight, but in Setting \ref{setting:alf}, the Riesz representer is generally signed depending on the linear functional $m$. In the current work we focus on the density ratio setting; however, the two settings are not mutually exclusive, as we illustrate through the causal inference examples below. In fact, many of the Riesz representer examples that motivated the consideration of Setting \ref{setting:alf}, can be viewed as density ratios.

\begin{example}[Outcome Regression]
\label{example:or}
To show the generality of the Riesz representer setup, consider Setting \ref{setting:alf}, with $m^{\text{(OR)}}(f, x, y) = f(x)y$. In this setting the Riesz representer is the outcome regression function $\alpha_0^{\text{(OR)}} = \mu_0$ (with $\mu_0(x)=\E_{P_0}[Y  \mid X=x]$ as above), since for all $f \in \mathcal{H}$
\begin{align*}
\E_{P_0}[m^{\text{(OR)}}(f, X, Y)] = \E_{P_0}[f(X)Y] = \langle f, \alpha_0^{\text{(OR)}} \rangle.
\end{align*}
Thus, one can view outcome regression as learning the Riesz representer for the bounded linear functional $f \mapsto \E[f(X) Y]$. In some sense, therefore, Riesz representer learning generalizes outcome regression to a broader class of statistical functionals.
\end{example}

\begin{example}[Average Treatment Effect]
\label{example:ate}
Let $X=(A, W)$ consist of a binary treatment $A \in \{0, 1\}$ and a vector of covariates $W\in \mathbb{R}^p$, and let $Y^a$ denote the outcome that would be observed if an intervention assigned treatment $A=a$. The average treatment effect $\E[Y^{1} - Y^{0}]$ is the difference in the mean outcome when treatment / no treatment is applied uniformly in the population \citep{Rosenbaum1983}.
Under standard causal assumptions, the average treatment effect is identified by $\E_{P_0}[m^{\text{(ATE)}}(\mu_0, X, Y)]$ where
$m^{\text{(ATE)}}(f, x, y) = f(1, w) - f(0, w)$
is a known linear functional. Hence, the average treatment effect is an example of Setting \ref{setting:alf}, and can be expressed as $\langle \mu_0, \alpha_0^{\text{(ATE)}} \rangle$ where
\begin{align*}
\alpha_0^{\text{(ATE)}}(x) \equiv \frac{\mathbb{I}\{a=1\}p_W(w) }{p_{A,W}(a,w)} - \frac{\mathbb{I}\{a=0\}p_W(w) }{p_{A,W}(a,w)}
\end{align*}
is a signed Riesz representer. The Riesz representer $\alpha_0^{\text{(ATE)}}$ also appears in pseudo-outcome learners for the conditional average treatment effect, such as in the DR-learner of \cite{kennedy_towards_2023}.
\end{example}

\begin{example}[Average Policy Effect]
\label{example:ape}
Using the setup from Example \ref{example:ate}, the average policy effect $\E[Y^{\pi(W)}]$ is the mean outcome when treatment is determined by a known decision rule $\pi: \mathbb{R}^p \to  \{0,1\}$ that is a function of covariates \citep{Dudik2011,VanderLaan2014,Athey2021}.
Under standard causal assumptions, the average policy effect is identified by $\langle \mu_0, \alpha_0^{\text{(APE)}} \rangle$ where
\begin{align*}
\alpha_0^{\text{(APE)}}(x) \equiv \frac{\mathbb{I}\{a=\pi(w)\}p_W(w) }{p_{A,W}(a,w)}
\end{align*}
is a density ratio (Riesz representer from Setting \ref{setting:dr}).
The average policy effect can also be expressed as $\E_{P_0}[m^{\text{(APE)}}(\mu_0, X, Y)]$ where
$m^{\text{(APE)}}(f, x, y) = \pi(w) f(1, w) + \{1-\pi(w)\}f(0, w)$
is a known linear functional for all $f(a, w)$ in $\mathcal{H}$. Hence, the average policy effect is also an example of Setting \ref{setting:alf}. Comparing the Riesz representer above with that in Example \ref{example:ate}, we see that $\alpha_0^{\text{(ATE)}}$ represents the difference in the density ratios corresponding to the average policy effects $\E[Y^{1}]$ and $\E[Y^{0}]$.
\end{example}

\begin{example}[Average Shift Effect]
\label{example:ase}
Let $X=(A, W)$ consist of a continuous treatment $A \in \mathbb{R}$ and covariates $W\in \mathbb{R}^p$.
Letting $Y^a$ denote the potential outcome, as in Example \ref{example:ape}, the average shift effect $\E[Y^{A + \delta}]$ is the mean outcome when the natural value of treatment is uniformly shifted by a constant $\delta \in \mathbb{R}$ \citep{diaz_population_2012,diaz_nonparametric_2023}.
Under standard causal assumptions, this quantity is identified by $\langle \mu_0, \alpha_0^{\text{(ASE)}} \rangle$ where
\begin{align*}
\alpha_0^{\text{(ASE)}}(x) \equiv \frac{p_{A,W}(a - \delta,w)}{p_{A,W}(a,w)}
\end{align*}
is a density ratio (Riesz representer from Setting \ref{setting:dr}).
The average shift effect can also be expressed as $\E_{P_0}[m^{\text{(ASE)}}(\mu_0, X, Y)]$ where
$m^{\text{(ASE)}}(f, x, y) = f(a+\delta, w)$
is a known linear functional for all $f(a, w)$ in $\mathcal{H}$. Hence, the average shift effect is also an example of Setting \ref{setting:alf}.
\end{example}

\begin{example}[Average Derivative Effect]
\label{example:ade}
Using the setup from Example \ref{example:ase}, the average derivative effect $\lim_{\delta \to 0} \E[Y^{A + \delta} - Y] / \delta$ is the limiting difference between the average shift effect and no intervention per unit change in treatment \citep{rothenhausler_incremental_2020}.
Under causal assumptions, and assuming that $\mu_0$ is differentiable with respect to treatment, the average derivative effect is identified by $\E_{P_0}[m^{\text{(ADE)}}(\mu_0, X, Y)]$,
where $m^{\text{(ADE)}}(f, x, y) = \partial_a f(a, x)$ and $\partial_a$ denotes the partial derivative with respect to $a$ \citep{hardle_investigating_1989,newey_efficiency_1993,imbens_identification_2009}.
Under regularity conditions regarding the boundary of the support of treatment, the average derivative effect is also identified by $\langle \mu_0, \alpha_0^{\text{(ADE)}} \rangle$, where
\begin{align*}
\alpha_0^{\text{(ADE)}}(x) \equiv - \partial_a \log\{p_{X}(a, w)\}
\end{align*}
is a Riesz representer from Setting \ref{setting:alf}. Although this Riesz representer is not a density ratio, it can be viewed as a limiting scaled difference between an average shift effect and no intervention (density ratio of one)
\begin{align*}
\alpha_0^{\text{(ADE)}}(x) = \lim_{\delta \to 0} \frac{1}{\delta}\left[ \frac{p_{A,W}(a - \delta,w)}{p_{A,W}(a,w)} - 1 \right].
\end{align*}
Moreover, in Appendix \ref{sect:score_matching} we show how a vectorized version of this Riesz representer forms the basis of score matching methods in generative diffusion modeling, and can be estimated by the Riesz representer learning methods in Section \ref{sect:breg} known as \textit{score matching} \citep{hyvarinen_estimation_2005,lyu_interpretation_2009,song_generative_2019}.
\end{example}

\begin{example}[Average Dose Response Curve]
\label{example:adrc}
Using the setup of Example \ref{example:ase}, the dose response curve $\psi(a) \equiv \E_{P_0}[Y^a]$ is the mean outcome when an intervention uniformly assigns treatment $A=a$ \citep{Robins2001,Kennedy2017}, and the average dose response curve $\E_{P_0}[\psi(A)]$ is its mean.
Under standard causal assumptions, the average dose response curve is identified by $\langle \mu_0, \alpha_0^{\text{(SW)}} \rangle$, where
\begin{align*}
\alpha_0^{\text{(SW)}}(x) \equiv \frac{p_A(a)p_W(w)}{p_{A,W}(a,w)}
\end{align*}
is a density ratio (Riesz representer from Setting \ref{setting:dr}), known as the stabilized weight \citep{robins_marginal_2000,ai_unified_2021}.
The stabilized weight also appears in pseudo-outcome based learners of the dose response curve itself \citep{Kennedy2017} and in the definition of the mutual information $-\E_{P_0}[\log\{\alpha_0^{\text{(SW)}}(X)\}]$ between $A$ and $W$ \citep{nguyen_estimating_2007,belghazi_mutual_2018}.
\end{example}

\begin{example}[Natural Mediation Effect]
\label{example:nme}
Let $X=(M, A, W)$ consist of a mediator $M\in \mathbb{R}^q$, a binary treatment $A \in \{0, 1\}$, and covariates $W\in \mathbb{R}^p$.
Also, let $Y^{a,m}$ denote the outcome that would be observed if treatment and mediator were set to $A=a$ and $M=m$, and let $M^a$ denote the mediator that would be observed if treatment were set to $A=a$. Natural mediation effects are defined via the quantity $\E[Y^{a^\prime, M^{a\dagger}}]$, where $a^\prime$ and $a^\dagger$ are two levels of treatment \citep{robins_identifiability_1992,pearl_direct_2001}.
Under causal assumptions given in \cite{pearl_direct_2001}, the natural mediation effect is identified by $\langle \mu_0, \alpha_0^{\text{(NME)}} \rangle$ where
\begin{align*}
\alpha_0^{\text{(NME)}}(x) &\equiv \frac{p_{M\mid A,W}(m\mid a^\dagger,w)\mathbb{I}\{a=a^\prime\}p_W(w) }{p_{M,A,W}(m,a,w)}
\end{align*}
is a density ratio (Riesz representer from Setting \ref{setting:dr}).
\end{example}

\begin{remark}
    The absolute continuity condition $P_1 \ll P_0$ in the density ratio setting (Setting \ref{setting:dr}) is also referred to as an overlap or positivity assumption. This condition can often be quite strong and unlikely to hold, especially when $\mathcal{X}$ is high-dimensional \citep{damour_overlap_2021}. In particular, even when population overlap is satisfied, practical overlap violations in finite samples may remain. That is, there may be regions of the covariate space $\mathcal{X}$ in which there are very few or no observations from $P_0$, but many observations from $P_1$. In the density ratio learning literature, this is also referred to as the density chasm problem \citep{rhodes_telescoping_2020}.
    See additional discussion on overlap in causal inference by \cite{Crump2009}, \cite{khan_irregular_2010}, \cite{petersen2012diagnosing}, and \cite{susmann_non-overlap_2025}.
\end{remark}

\begin{remark}
\label{remark:point_estimators}
One of the key reasons for Riesz representer learning in causal inference is to obtain efficient and debiased point estimates for the quantity $\h(\mu_0)$ in the average linear functional setting (Setting \ref{setting:alf}). Here we give a very brief account of this theory and refer to \cite{rotnitzky_characterization_2021} and \cite{Hirshberg2021} for more details.
Given a point-wise consistent regression estimator $\hat{\mu}$ for $\mu_0$, a naive estimator for $\h(\mu_0)$ can be constructed as $\h_n(\hat{\mu})$, where $\h_n(f) \equiv \E_n[m(f, X, Y)]$ and $\E_{n}[\cdot] \equiv n^{-1}\sum_{i=1}^n [\cdot]_i$ denotes expectation under an empirical measure given $n$ samples from $P_0$. However, this naive estimator is generally biased since, under limited regularity conditions (e.g., when $\hat{\mu}$ is learned on an independent sample) one obtains the asymptotic result
\begin{align*}
    \sqrt{n} \left\{ \h_n(\hat{\mu})- \h(\mu_0) \right\} + \sqrt{n}\E_{n}[\alpha_0(X)\{Y-\hat{\mu}(X)\}] \overset{d}{\to} \n\left(0, v_0 \right),
\end{align*}
where $v_0 = \Var_{P_0}[m(\mu_0, Y, X) + \alpha_0(X)\{Y-\mu_0(X)\}]$. In this result, the term $\sqrt{n}\E_{n}[\alpha_0(X)\{Y-\hat{\mu}(X)\}]$ generally does not converge to zero and therefore represents an asymptotic bias in the naive estimator. This bias can be estimated as $\E_{n}[\hat{\alpha}(X)\{Y-\hat{\mu}(X)\}]$ using a learner $\hat{\alpha}$ of the Riesz representer, and can be viewed as an importance weighted estimate of $\h(\mu_0 - \hat{\mu}) = \E_{P_0}[\alpha_0(X)\{Y - \hat{\mu}(X)\}]$. Estimators which correct for this bias are referred to as debiased (because they correct for the bias), efficient (because they have variance $v_0$ equal to the semiparametric efficiency bound), and rate double robust (because they require a mixed bias condition $\sqrt{n}\langle \hat{\mu} - \mu_0, \hat{\alpha} - \alpha_0\rangle \to 0$, which allows for a trade-off in the learning rates of $\hat{\mu}$ and $\hat{\alpha}$).
\end{remark}

\subsection{Bregman--Riesz regression}
\label{sect:breg}

Here we outline Bregman--Riesz regression as any method that minimizes an empirical approximation of a Bregman--Riesz Risk (BRR), which we define presently.
The BRR generalizes the Bregman divergence frameworks from the density ratio learning methods of \cite{nguyen_estimating_2010}, \cite{gutmann_bregman_2011}, and \cite{sugiyama_density-ratio_2012} to Riesz representer learning.
Bregman divergences \citep{bregman_relaxation_1967,lafferty_additive_1999,grunwald_game_2004} are a fundamental class of convex divergences that generalize the squared Euclidean distance.
Letting $F:\mathbb{R} \to \mathbb{R}$ be a known, strictly convex function with derivative $F^\prime$, the unit Bregman divergence is defined for any pair of values $v_0, v_1$ on the domain of $F$ as
\begin{align}
    d_F(v_0 \| v_1) &\equiv F(v_0) - F(v_1) - F^\prime(v_1)(v_0 - v_1). \label{defn:breg}
\end{align}
This divergence can be interpreted as the residual of the Taylor first-order approximation $F(v_0) \approx F(v_1) + F^\prime(v_1)(v_0 - v_1)$. Due to the convexity of $F$, $d_F(v_0 \| v_1)$ is convex in $v_0$ and $d_F(v_0\|v_1) \geq 0$ with equality if and only if $v_0=v_1$.
This motivates characterizing the unknown Riesz representer $\alpha_0$ as a population risk minimizer with the unit Bregman divergence as a unit loss. That is, for $\alpha \in \mathcal{H}$ we define the risk
\begin{align*}
    D_F(\alpha_0 \| \alpha) \equiv \E_{P_0}[d_F\{\alpha_0(X)\| \alpha(X)\}]
\end{align*}
which we call the Bregman divergence.
This divergence has the property that $D_F(\alpha_0 \| \alpha) \geq 0$ with equality if and only if $\alpha = \alpha_0$ almost surely.
Our estimator $\hat{\alpha}$ will minimize an empirical approximation of $D_F(\alpha_0 \| \alpha)$, over functions $\alpha$, ignoring additive constants that do not affect the minimization problem.
% Note that the Bregman divergence and its mean can both be viewed as special cases of the pathwise form $f(0) - f(1) + f^\prime(1)$ for $f(t) = F\{(1-t)v_0 + tv_1\}$ and $f(t) = \E_{P_0}[F\{(1-t)g_0(X) + t g_1(X)\}]$ respectively.
% This makes Bregman divergences useful in machine learning and in optimization problems based on empirical risk minimization.

To gain some intuition for the choice of convex generating function $F$, consider the case where the second derivative $F^{\prime\prime}$ exists.
The Bregman divergence can be expressed in terms of the \textit{cost-sensitive point-wise regret}
\begin{align*}
    r(t \mid v_0, v_1) &\equiv \lvert t - v_0 \rvert \mathbb{I}\left\{ \min(v_0, v_1) < t < \max(v_0, v_1) \right\},
\end{align*}
which, for a true value $v_0 \in \mathbb{R}$ and an estimate $v_1 \in \mathbb{R}$, returns the absolute error of an alternative estimate $t \in \mathbb{R}$ when $t$ is better than $v_1$, in the sense of being between $v_1$ and $v_0$, and returns zero otherwise.
The Bregman divergence is the integral of the mean regret over Riesz representer values with value weights $F^{\prime\prime}$
\begin{align}
    D_F(\alpha_0 \| \alpha) &= \int_{-\infty}^\infty \E_{P_0}\left[ r\{t\mid \alpha_0(X), \alpha(X)\} \right] F^{\prime\prime}(t)\dd t. \label{eq:bman_cost}
\end{align}
See Appendix \ref{sect:breg_proof} for details and \cite{reid_surrogate_2009} for related discussion.
This integral form suggests that the divergence $D_F(\alpha_0\|\alpha)$ gives greater emphasis towards accurately modeling $\alpha_0(X)$ for values where $F^{\prime\prime}\{\alpha_0(X)\}$ is large.
In this way, the integral weight $F^{\prime\prime}(t)$ can be viewed as an analog of the variance function $V(t) = 1 / F^{\prime\prime}(t)$ in the theory of generalized linear models, see Appendix \ref{sect:glm_deviance}.
%Additionally, note that the Bregman divergence is invariant to the addition of an affine function to the convex generating function $F$, since such modifications do not affect $F^{\prime\prime}$.

One of the key advantages of basing Riesz representer learning around Bregman divergences is that the divergence $D_F(\alpha_0 \| \alpha)$ implies a risk $\mathcal{R}_F(\alpha)$ %, which we call the Bregman--Riesz Risk (BRR),
that depends only implicitly on the unknown Riesz representer through the bounded linear functional $\h$. We refer to this risk as the \emph{Bregman--Riesz Risk (BRR)}, which is defined as
\begin{align}
    \mathcal{R}_F(\alpha) &\equiv \E_{P_0}\left[F^\prime\{\alpha(X)\} \alpha(X) - F\{\alpha(X)\} \right] - \h\left(F^\prime \circ \alpha \right), \label{eq:bman}
\end{align}
and is minimized at $\alpha_0$.
For proof, note that
\begin{align*}
    \mathcal{R}_F(\alpha) &= \E_{P_0}\left[- F\{\alpha(X)\} - F^\prime\{\alpha(X)\} \{\alpha_0(X) - \alpha(X)\}\right]
\end{align*}
and the Bregman divergence is the excess risk $D_F(\alpha_0 \|\alpha) = \mathcal{R}_F(\alpha) - \mathcal{R}_F(\alpha_0)$.
Discarding the constant $\mathcal{R}_F(\alpha_0)$ it follows that
\begin{align*}
    \alpha_0 &= \argmin_{\alpha \in \mathcal{H}} D_F(\alpha_0 \|\alpha)  \\
    &= \argmin_{\alpha \in \mathcal{H}} \mathcal{R}_F(\alpha).
\end{align*}
In the density ratio and average linear functional settings (Settings \ref{setting:dr} and \ref{setting:alf}), the BRR, respectively, reduces to
\begin{align}
\mathcal{R}_F(\alpha) &= \E_{P_0}\left[F^\prime\{\alpha(X)\} \alpha(X) - F\{\alpha(X)\} \right] - \E_{P_1} \left[F^\prime\{\alpha(X)\} \right], \label{eq:breg_dr}\\
\mathcal{R}_F(\alpha) &= \E_{P_0}\left[F^\prime\{\alpha(X)\} \alpha(X) - F\{\alpha(X)\} - m\left(F^\prime \circ \alpha, X, Y\right) \right]. \label{eq:breg_alf}
\end{align}
Neither of these expressions depend on the unknown $\alpha_0$, making them useful for learning $\alpha_0$ via empirical risk minimization.
We show the form of the BRR for various choices of convex differentiable functions $F$ in Table \ref{tab:breg}. For Bregman divergences using other functions, see \citet[Table 1]{menon_linking_2016} and \citet[Table 1]{banerjee_clustering_2004}.

\begin{table}[htbp]
\caption{Summary of Bregman--Riesz Risks considered in this paper, as defined by their generating function $F$ which is convex on the given domain. The second derivative $F^{\prime\prime}$ relates to the cost weight in the integral Bregman divergence expression in \eqref{eq:bman_cost}.}
\label{tab:breg}
\centering
\begin{tabular}{llll}
\hline\hline
Name & Domain &$F(t)$ & $F^{\prime\prime}(t)$ \\ \hline
Least squares & $\mathbb{R}$ & $t^2/2$ & $1$ \\
Kullback--Leibler & $\mathbb{R}_{>0}$ & $t \log(t) - t$ &  $t^{-1}$\\
Negative binomial & $\mathbb{R}_{>0}$ & $t \log(t) - (1 + t) \log(1 + t)$ & $t^{-1}(1+t)^{-1}$\\
Itakura--Saito & $\mathbb{R}_{>0}$ & $- \log(t) - 1$ &  $t^{-2}$ \\
\hline
\end{tabular}
\end{table}

\begin{divergence}[Least squares]
For $F(t) = t^2/2$ the divergence in \eqref{defn:breg} becomes the mean squared error $D_F(\alpha_0 \|\alpha) = \E_{P_0}[ \{\alpha_0(X) - \alpha(X) \}^2] /2$ and the BRR reduces to
\begin{align*}
    \mathcal{R}_F(\alpha) &= \frac{1}{2}\E_{P_0}\left[\alpha^2(X) \right] - \h\left(\alpha\right).
\end{align*}
The least-squares divergence was studied in the density ratio setting by \cite{kanamori_least-squares_2009}, who propose Least-Squares Importance Fitting (LSIF) and unconstrained LSIF (uLSIF) procedures based on empirical minimization of the least-squares BRR when $\alpha(x)$ is a sieve model, see Section \ref{sect:simulations}. In the average linear functional setting, \cite{chernozhukov_automatic_2024} propose neural network Riesz representer learners based on empirical minimization of the least-squares BRR in a procedure referred to as Riesz regression. An earlier special case of Riesz regression is the score matching proposal of \cite{hyvarinen_estimation_2005}, see Appendix \ref{sect:score_matching}.
\end{divergence}

\begin{divergence}[Kullback--Leibler]
\label{div:kl}
When $\alpha_0(X) > 0$ and $\alpha(X) > 0$ almost surely then one can let $F(t) = t \log(t) - t$, in which case the divergence in \eqref{defn:breg} is the Kullback--Leibler (KL) divergence
\begin{align*}
    D_F(\alpha_0 \|\alpha) &= \E_{P_0}\left[\alpha_0(X) \log\left\{\frac{\alpha_0(X)}{\alpha(X)}\right\} +\alpha(X) - \alpha_0(X) \right].
\end{align*}
This version of the KL divergence is \textit{unnormalized}, with the more conventional normalized version obtained when $\E_{P_0}\left[\alpha(X)\right] =  \E_{P_0}\left[\alpha_0(X) \right]$ and hence the term $\E_{P_0}\left[\alpha(X) - \alpha_0(X) \right]$ is zero. Note that, in the density ratio setting with denominator density $p_0(x)$, $\E_{P_0}\left[\alpha_0(X) \right] = 1$ a priori because the numerator density integrates to 1. The BRR corresponding to the KL divergence is
\begin{align*}
    \mathcal{R}_F(\alpha) &= \E_{P_0}\left[\alpha(X) \right] - \h\left(\log \circ \alpha \right).
\end{align*}
 This KL BRR was studied in the density ratio setting by \cite{sugiyama_direct_2007}, who propose the KL Importance Estimation Procedure (KLIEP) based on minimizing an empirical approximation of $-\h\left(\log \circ \alpha \right)$ subject to the constraint that, empirically, $\E_{P_0}\left[\alpha(X) \right] = 1$ and $\alpha(x)$ is a sieve model, see Section \ref{sect:simulations}. Their original derivation considers the KL divergence between the numerator density $p_1(x) = \alpha_0(x) p_0(x)$ and $\tilde{p}_1(x) \equiv \alpha(x) p_0(x)$, which is a density when $\E_{P_0}[\alpha(X)] = 1$.
\end{divergence}

\begin{divergence}[Negative binomial]
When $\alpha_0(X) > 0$ and $\alpha(X) > 0$ almost surely one can let $F(t) = t \log(t) - (1 + t) \log(1 + t)$ in which case the divergence in \eqref{defn:breg} and BRR are
\begin{align*}
    D_F(\alpha_0 \|\alpha) &= \E_{P_0}\left[\alpha_0(X) \log\left\{\frac{1 + 1/\alpha(X)}{1 + 1/\alpha_0(X)}\right\} +\log\left\{\frac{1 + \alpha(X)}{1 + \alpha_0(X)}\right\} \right] \\
    \mathcal{R}_F(\alpha) &=  \E_{P_0}[\log\{1 + \alpha(X)\}] + \h\left[\log\{1 + 1/\alpha(\cdot)\}\right].
\end{align*}
In Section \ref{sect:prob} we show how, in the density ratio setting, the negative binomial divergence is connected to the binary cross-entropy loss used in classification. We call this divergence `negative binomial' due to its connection with the negative binomial generalized linear model deviance, see Appendix \ref{sect:glm_deviance}.
The connection between density ratio learning and probabilistic classification has previously been established by \cite{qin_inferences_1998} for parametric models, and \cite{menon_linking_2016} for Bregman divergences.
The generating function $F$ for the negative binomial divergence has the property $F(t) = tF(1/t)$, which we argue relates to a kind of symmetry in the density ratio setting when the roles of $P_0$ and $P_1$ are reversed, see Appendix \ref{sect:sym}.
\end{divergence}

\begin{divergence}[Itakura--Saito]
When $\alpha_0(X) > 0$ and $\alpha(X) > 0$ almost surely, one can let $F(t) = -\log(t) - 1$, in which case the divergence in \eqref{defn:breg} and BRR are
\begin{align*}
    D_F(\alpha_0 \|\alpha) &= \E_{P_0}\left[\frac{\alpha_0(X)}{\alpha(X)} - \log\left\{\frac{\alpha_0(X)}{\alpha(X)} \right\} - 1\right] \\
    \mathcal{R}_F(\alpha) &= \E_{P_0}\left[\log\left\{\alpha(X)\right\}\right]  + \h\left[\frac{1}{\alpha(\cdot)}\right].
\end{align*}
This divergence is known as the Itakura--Saito distance and has previously been used for matrix factorization and audio analysis \citep{banerjee_clustering_2004}. In the density ratio setting, the Itakura--Saito distance is equivalent to the unnormalized Kullback--Leibler divergence (Divergence \ref{div:kl}) when the roles of $P_0$ and $P_1$ are reversed, see Appendix \ref{sect:sym}. To our knowledge, this divergence has not previously been used for density ratio learning, but we include it because of its connection with the Kullback--Leibler divergence.
\end{divergence}

\begin{remark}
Viewing the BRR in \eqref{eq:breg_alf} as the mean of a unit loss, the unit loss generally depends on the value of $\alpha$ at multiple points of the input space $\mathcal{X}$ through the term $m\left(F^\prime \circ \alpha, x, y\right)$. For example, for the average treatment effect (Example \ref{example:ate}), the unit loss in \eqref{eq:breg_alf} is
\begin{align*}
F^\prime\{\alpha(a, w)\} \alpha(a, w) - F\{\alpha(a, w)\} - F^\prime\{\alpha(1, w)\} + F^\prime\{\alpha(0, w)\},
\end{align*}
which depends on the values $\alpha(1, w)$ and $\alpha(0, w)$. This makes Riesz representer learning via empirical loss minimization challenging for some algorithms, such as gradient boosting \citep{Lee2025}.
\end{remark}

\begin{remark}
\label{remark:duality}
Letting $\beta(x) = F^\prime\{\alpha(x)\}$, the BRR can be written as
\begin{align*}
    \mathcal{R}_F(\alpha) &= \E_{P_0}\left[F^*\{\beta(X)\} \right] - \h\left(\beta \right)
\end{align*}
where $F^*(u) \equiv \sup_t ut-F(t)$ is the convex conjugate of $F$. The function $F^\prime$ is an analog of the canonical link function from the standard theory for generalized linear models, where $\beta(x)$ is modeled as a linear predictor, see Appendix \ref{sect:glm_deviance}.
\end{remark}

\begin{remark}
    In the analysis of estimators that use Riesz representer estimates, such as those in Remark \ref{remark:point_estimators}, it is often of interest to bound the error $\|\hat{\alpha} - \alpha_0\|_{\mathcal{H}}$.
    This is controlled by the Bregman divergence under strong convexity, which is said to hold when there exists a constant $k_F$ such that $k_F\|\hat{\alpha} - \alpha_0\|_{\mathcal{H}}^2 \leq D_F(\alpha_0\|\hat{\alpha})$. In particular, $k_F = 2 \inf_{t\in[a, b]} F^{\prime\prime}(t)$ is such a constant when $F^{\prime\prime}$ exists and both $\hat{\alpha}(X)$ and $\alpha_0(X)$ lie on an interval $[a, b]$ with probability 1, see Appendix \ref{sect:breg_proof}.
\end{remark}

\begin{remark}
    \label{remark:ate}
    The Itakura--Saito distance with $F^{\prime\prime}(t) = t^{-2}$ implies a generating function $F(t) = -\log(|t|) - 1$ which is strictly convex on the intervals $(-\infty, 0)$ and $(0, \infty)$. The Itakura--Saito distance can therefore support negative Riesz representer values by taking the absolute value of the inputs to each logarithm. Maintaining convexity requires that $\alpha_0(X)\alpha(X) > 0$ almost surely, i.e. $\alpha_0(X)$ and $\alpha(X)$ are non zero with the same sign. This requirement is easily satisfied for the average treatment effect (Example \ref{example:ate}), since the sign of $\alpha_0^{(ATE)}$ is known and fully determined by the binary treatment. The BRR expression in \eqref{eq:breg_alf} with $m = m^{\text{(ATE)}}$ and $F(t) = -\log(|t|) - 1$ becomes
    \begin{align*}
    \mathcal{R}_F(\alpha) = \E_{P_0}\left[ \log\left\{ |\alpha(A, W)|\right\} + \frac{1}{\alpha(1, W)} - \frac{1}{\alpha(0, W)}\right].
    \end{align*}
    If one restricts $\alpha$ to have the same functional form as $\alpha_0^{(ATE)}$, with a propensity score $p: \mathbb{R}^p \to  (0, 1)$, this BRR reduces to
    \begin{align*}
    \alpha(a, w) &= \frac{a}{p(w)} - \frac{1 - a}{1-p(w)} \\
    \implies \mathcal{R}_F(\alpha) &= \E_{P_0}\left[ -A\log\{p(W)\} - (1-A)\log\{1-p(W)\}\right] + 1,
    \end{align*}
    which, ignoring the plus one, is the binary cross-entropy for learning $A$ given $W$. Hence, the following procedures are equivalent for learning the Riesz representer of the average treatment effect: (i) minimize an empirical estimate of the Itakura--Saito distance while imposing the known functional form of the Riesz representer; and (ii) learn the propensity by minimizing an empirical estimate of the binary cross-entropy and then construct Riesz representer estimates using the known functional form.
\end{remark}

\subsection{Density ratio learning using Bregman--Riesz regression}
\label{sect:dr}

Here we restrict attention to the setting where the target Riesz representer is a density ratio. Typically, the density ratio learning literature assumes a standard setup where one has access to $n_1$ samples from $P_1$ and $n_0$ samples from $P_0$.
However, in causal examples, an observation can often be viewed as a sample from both the numerator and denominator distributions, e.g. if the observed treatment happens to equal the intervened treatment under a single time point intervention. Therefore, it is more practical to consider a slightly more general setup. We presume access to samples $\{(x_i, \gamma^0_i, \gamma^1_i)\}_{i=1}^n$ consisting of predictors and nonnegative weights such that $x_i$ weighted by $\gamma^0_i$ represents a sample from $P_0$ and $x_i$ weighted by $\gamma^1_i$ represents a sample from $P_1$, i.e., for a test function $f\in\mathcal{H}$
\begin{align*}
    \frac{1}{n}\sum_{i=1}^n \gamma^0_i f(x_i) &\approx \E_{P_0}[f(X)] \\
    \frac{1}{n}\sum_{i=1}^n \gamma^1_i f(x_i) &\approx \E_{P_1}[f(X)] = \h(f).
\end{align*}
In Section \ref{sect:augment} we show how data augmentation techniques can be used in practice to obtain such samples for several causal problems. The standard setup, where one has access to $n_1$ samples from $P_1$ and $n_0$ samples from $P_0$, can be expressed in this framework by letting $n = n_0 + n_1$, and letting the index $i$ iterate through the samples from $P_0$ with $(\gamma^0_i, \gamma^1_i) = (n/n_0, 0)$ followed by the samples from $P_1$ with $(\gamma^0_i, \gamma^1_i) = (0, n/n_1)$. Supposing that it is possible to obtain such samples, the BRR can be empirically estimated as
\begin{align}
    \mathcal{R}_{F,n}(\alpha) &\equiv \frac{1}{n} \sum_{i=1}^n - \gamma^0_i F\{\alpha(x_i)\} - F^\prime\{\alpha(x_i)\}\{\gamma^1_i - \gamma^0_i\alpha(x_i)\}. \label{eq:rn_normal}
\end{align}
The resulting density ratio learning algorithm is summarized in Algorithm \ref{alg:dr}. This algorithm is agnostic to the form of the model class $\mathcal{M}$ and regularizer $\Lambda_n$ which can represent one of a broad class of machine learning algorithms (neural networks, gradient boosted trees, lasso, etc.).
In Section \ref{sect:prob} we outline how Step 3 of Algorithm \ref{alg:dr} can be implemented in practice using existing software tools for outcome regression.

\begin{algorithm}[Bregman--Riesz regression for density ratio learning]
\label{alg:dr}\hfill
\begin{enumerate}
    \item Choose a convex generating function $F$.
    \item Use an augmentation strategy to obtain samples and weights $\{(x_i, \gamma^0_i, \gamma^1_i)\}_{i=1}^{n}$.
    \item Learn the density ratio as $\hat{\alpha} = \argmin_{\alpha \in \mathcal{M}} \left\{ \mathcal{R}_{F,n}(\alpha) + \Lambda_n(\alpha) \right\}$, where $\mathcal{R}_{F,n}(\alpha)$ is defined in \eqref{eq:rn_normal}, $\mathcal{M}$ is a user-defined function class, and $\Lambda_n: \mathcal{M} \to \mathbb{R}$ is an optional penalization term.
\end{enumerate}
\end{algorithm}

\begin{remark}
    We can compare the empirical BRR estimator in \eqref{eq:rn_normal}, for learning density ratios, with the empirical BRR estimator for outcome regression, as in Example \ref{example:or}. Given observations $\{(x_i, y_i)\}_{i=1}^n$ from $P_0$, an outcome regression BRR estimator is obtained by letting $(\gamma_i^0, \gamma_i^1) = (1, y_i)$ in \eqref{eq:rn_normal}, resulting in
    \begin{align}
        \mathcal{R}_{F,n}(\mu) &\equiv \frac{1}{n} \sum_{i=1}^n - F\{\mu(x_i)\} - F^\prime\{\mu(x_i)\}\{y_i - \mu(x_i)\}. \label{eq:rn_outcome}
    \end{align}
    This substitution works because, for a test function $f$, $n^{-1}\sum_{i=1}^n f(x_i)$ is an estimator for $\E_{P_0}[f(X)]$ and $n^{-1}\sum_{i=1}^n y_i f(x_i)$ is an estimator for $\h(f) = \langle\mu_0, f\rangle = \E_{P_0}[Yf(X)]$.
\end{remark}

\subsection{Data augmentation}
\label{sect:augment}

Here we show how data augmentation methods, similar to those of \cite{Lee2025} and \cite{diaz_nonparametric_2023}, can be used in Step 2 of Algorithm \ref{alg:dr} to obtain samples and weights $\mathcal{D}_n = \{(x_i, \gamma^0_i, \gamma^1_i)\}_{i=1}^{n}$. The standard density ratio learning setup supposes access to $n_1$ samples from $P_1$ and $n_0$ samples from $P_0$, resulting in a combined sample of $n = n_0 + n_1$ observations. However, in the causal inference examples from Section \ref{sect:setup}, $P_1$ represents a counterfactual intervention distribution from which one does not have samples and $P_0$ represents the observed data distribution from which one does have samples. Therefore, to approximate expectations of the form $\E_{P_1}[f(X)]$, additional techniques are required.

We view the requirement for data augmentation as one of the main practical barriers to using density ratio learning methods in causal inference. We address this issue by demonstrating possible augmentation schemes in several examples. These examples use a combination of three main strategies: (a) decomposing the density ratio learning problem into a series of simpler learning problems, (b) augmenting the observed data, and (c) reweighting the augmented data.

To ease notation, we let $\mathcal{D}_n = \{(x_i, \tilde{\gamma}^0_i, \tilde{\gamma}^1_i)\}_{i=1}^n$ denote the target sequence of samples with unnormalized weights that are normalized as, e.g. $\gamma^0_i = \tilde{\gamma}_i^0/ n^{-1}\sum_{j=1}^n \tilde{\gamma}_j^0$, with similar for $\gamma_i^1$. We also let the symbol $\frown$ denote the concatenation of two sequences. For example, the standard density ratio setup, where one combines $n_0$ samples from $P_0$ and $n_1$ samples from $P_1$, can be expressed as $\mathcal{D}_n =  \{(x_i, 1, 0)\}_{i=1}^{n_0} \frown \{(x_i, 0, 1)\}_{i=1}^{n_1}$. In each example below, we suppose access to $n_0$ samples $\{x_i\}_{i=1}^{n_0}$ from $P_0$.

\begin{augmentation_example}[Modified Treatment Policy]
\label{augment:mtp}
In Examples \ref{example:ape} and \ref{example:ase}, the intervention assigns a new treatment $\tilde{a}_i = \tilde{a}(a_i, w_i)$ that depends on the natural value of treatment $a_i$, and/or covariates $w_i$ \citep{diaz_population_2012,haneuse_estimation_2013}.
Specifically, for the average policy effect $\tilde{a}_i = \pi(w_i)$ and for the average shift effect $\tilde{a}_i = a_i + \delta$.
Thus, $n_0$ observations from $P_1$ can be constructed as $\{(\tilde{a}_i, w_i)\}_{i=1}^{n_0}$. Combining these with the observed data results in the augmented sample $\mathcal{D}_n =  \{((a_i, w_i), 1, 0)\}_{i=1}^{n_0} \frown \{((\tilde{a}_i, w_i), 0, 1)\}_{i=1}^{n_0}$.
This technique is used by \cite{Lee2025} for the average policy/shift effect, and by \cite{diaz_nonparametric_2023} for the average shift effect and for similar modified treatment policy interventions, generalizing to the setting of longitudinal treatments.
\end{augmentation_example}

\begin{augmentation_example}[Binary Modified Treatment Policy]
\label{augment:btp}
In Example \ref{example:ape}, the density ratio $\alpha^{(APE)}$ is zero unless $a = \pi(w)$. For this reason, Bregman divergences that are defined only on positive values may not be appropriate for learning $\alpha^{(APE)}$ directly using the strategy in Augmentation Example \ref{augment:mtp}.
However, one can factorize the density ratio as
\begin{align*}
    \alpha^{(APE)}(x) = \frac{\mathbb{I}\{a = \pi(w)\}}{p_{A}\{\pi(w)\}} \cdot \frac{p_W(w)}{p_{W\mid A}\{w\mid \pi(w)\}}
\end{align*}
where the first term is known up to a normalizing constant $p_{A}\{\pi(w)\} = \E_{P_0}[\mathbb{I}\{A = \pi(W)\}]$ that is easy to estimate.
This factorization suggests learning $\alpha^{(APE)}$ by estimating this normalization constant, and separately learning the density ratio $p_W(w) / p_{W\mid A}\{w\mid \pi(w)\}$.
This can be achieved via Bregman--Riesz regression with $\mathcal{D}_n = \{(w_i, \tilde{\gamma}_i^0, 1)\}_{i=1}^{n_0}$ and $\tilde{\gamma}_i^0 = \mathbb{I}(a_i = \pi(w_i))$.
That is, to estimate empirical means over the numerator density one can average over all samples, and to estimate empirical means over the denominator density one can average over the samples where $a_i = \pi(w_i)$.
\end{augmentation_example}

\begin{augmentation_example}[Average Treatment Effect]
\label{augment:ate}
In Example \ref{example:ate}, the Riesz representer $\alpha^{(ATE)}$ is not a density ratio, but does admit a similar factorization to that in Augmentation Example \ref{augment:btp}
\begin{align*}
\alpha_0^{\text{(ATE)}}(x) = \left\{ \frac{\mathbb{I}\{a=1\}}{p_{A}(1)} - \frac{\mathbb{I}\{a=0\}}{p_{A}(0)} \right\} \cdot \frac{p_A(a)p_W(w)}{p_{A,W}(a,w)}.
\end{align*}
The second term is the stabilized weight $\alpha^{(SW)}$ of Example \ref{example:adrc}, except with a binary treatment $A$.
Similar factorizations have also been considered in work by \cite{arbour_permutation_2021} and \cite{sondhi_balanced_2020} on \textit{permutation weighting}, which use probabilistic classification to learn stabilized weights. The decomposition above suggests a procedure where one first estimates the constant $p_{A}(1)$ as $\hat{p}_1$, with $p_A(0)$ estimated by $1 - \hat{p}_1$, and then separately learns $\alpha^{(SW)}$. To do so, an augmented data set $\mathcal{D}_n = \{((1, w_i), a_i, \hat{p}_1)\}_{i=1}^{n_0} \frown \{((0, w_i), 1 - a_i, 1 - \hat{p}_1)\}_{i=1}^{n_0}$ can be constructed, where each covariate vector $w_i$ appears twice, once with/without treatment.
\end{augmentation_example}

\begin{augmentation_example}[Stabilized Weight]
\label{augment:sw}
Augmentation Example \ref{augment:ate} showed how the stabilized weight can be estimated when there are only two levels of treatment, resulting in an augmented dataset with two copies of the data. However, when treatment is continuous, an alternative strategy is required. First note that $(a_i, w_j)$, for a pair of indices $(i,j)$, represents a random draw from $p_A(a) p_W(w)$. Thus, one could consider the augmented dataset $\mathcal{D}_n = \{(a_i, w_j), \delta_{i,j}, 1\}_{(i,j) \in I}$, where the index set $I = \{1, ..., n_0\}^2$ consists of all pairs of observations, and $\delta_{ij}$ is a Kronecker delta, i.e. $\delta_{ij} = 1$ if $i=j$ and $0$ otherwise.

However, the resulting dataset, of size $n = n_0^2$, may be impractical even for modest $n_0$ due to long training times or computationally memory costs. Additionally, even if it were feasible to use all possible pairs, it may be preferable to ignore samples where $i = j$ from $P_1$, resulting in $\mathcal{D}_n = \{(a_i, w_j), \delta_{i,j}, 1 - \delta_{ij}\}_{(i,j) \in I}$. This distinction is equivalent to that of estimating $\E_{P_1}[\cdot]$ using a V-statistic versus a U-statistic, with the latter being the minimum variance unbiased estimator \citep[Chapter 12]{van_der_vaart_asymptotic_1998}, \citep{zhou_v_2021}.

In order to reduce the size $n$ of the augmented data, we instead consider drawing samples $\{(\tilde{a}_i, \tilde{w}_i)\}_{i=1}^{n_1}$ from $P_1$ and constructing $\mathcal{D}_n = \{((a_i, w_i), 1, 0)\}_{i=1}^{n_0} \frown \{((\tilde{a}_i, \tilde{w}_i), 0, 1)\}_{i=1}^{n_1}$. We consider the following schemes for drawing samples from $P_1$.
\begin{itemize}
\item \textbf{Sampling with (without) replacement}: $\{\tilde{a}\}_{i=1}^{n_1}$ is obtained by sampling from $\{a_i\}_{i=1}^{n_0}$ with (without) replacement and, $\{\tilde{w}\}_{i=1}^{n_1}$ is obtained by sampling from $\{w_i\}_{i=1}^{n_0}$ with (without) replacement and independently of $\{\tilde{a}\}_{i=1}^{n_1}$.

\item \textbf{Permutation}: Sample without replacement $n_1 = n_0$ times. Ignoring the order or the observations, this results in a sample $\{(\tilde{a}_i, w_i)\}_{i=1}^{n_1}$ where $\{\tilde{a}\}_{i=1}^{n_1}$ is a random permutation of $\{a_i\}_{i=1}^{n_0}$.

\item \textbf{Derangement}: A permutation is used that has no fixed points, also called a derangement.

\item \textbf{m-Permutations}/\textbf{m-Derangements}: The permutation/derangement method is repeated $m$ times to obtain $n_1 = m n_0$ draws.

\item \textbf{Train-time Sampling/Permutation/Derangement}: For learning algorithms based on stochastic gradient descent, the sampling/permutation/derangement methods can be repeated for each learning iteration or batch sample. See, e.g., \citet[Algorithm 1]{belghazi_mutual_2018} for a similar proposal where minibatches are permuted in neural network training.
\end{itemize}
Given the absence of clear practical recommendations regarding the sampling scheme for stabilized weight learning, we compare several methods through a numerical study in Section \ref{sect:simulations}. Specifically, we compare sampling with replacement ($n_1 = m n_0$) versus m-permutations and m-derangements for $m \in \{1, 2, 5, 10\}$. For neural network based models we also compare train-time sampling with replacement/permutation/derangement, each with $n_1 = n_0$.
\end{augmentation_example}

\begin{augmentation_example}[Natural Mediation Effect]
Consider the density ratio $\alpha_0^{\text{(NME)}}(x)$ from Example \ref{example:nme}.
Other proposals for learning $\alpha_0^{\text{(NME)}}(x)$, such as those of \cite{wu_density_2024}, are based on decomposing the density ratio into products of odds functions from probabilistic classifiers.
Instead, we decompose the density ratio as
\begin{align*}
\alpha_0^{\text{(NME)}}(x) &= \frac{\mathbb{I}\{a=a^\prime\}}{p_A(a^\prime)} \cdot \frac{p_W(w)}{p_{W\mid A}(w\mid a^\dagger)}  \cdot \frac{p_{M,W\mid A}(m,w\mid a^\dagger)}{p_{M,W|A}(m, w\mid a^\prime)}.
\end{align*}
which suggests estimating the constant $p_A(a^\prime)$ required for the first term and then learning the second and third terms separately using the strategy in Augmentation Example \ref{augment:btp}. However, the resulting learner for $\alpha_0^{\text{(NME)}}$ would represent a product of two separate density ratio learners, which may be undesirable since it does not control the Bregman divergence of the resulting $\alpha_0^{\text{(NME)}}$ learner directly, see \cite{Kunzel2019} for a discussion of similar metalearning issues.
Instead, estimates of the second term $\hat{\alpha}^{(2)}(w_i)$ could be used as weights, when learning the third term, resulting in a two-stage learner for the product of the second and third terms. This could be achieved using the augmented data set $\mathcal{D}_n = \{((m_i, w_i), \tilde{\gamma}_i^0, \tilde{\gamma}_i^1)\}_{i=1}^{n_0}$ with $\tilde{\gamma}_i^0 = \mathbb{I}(a_i = a^\prime)$ and $\tilde{\gamma}_i^1 = \hat{\alpha}^{(2)}(w_i)\mathbb{I}(a_i = a^\dagger)$.
\end{augmentation_example}

\begin{remark}
    The problem of drawing samples from an intervention distribution $P_1$ is closely related to the problem of drawing counterfactual outcomes under an intervention. The latter problem is of interest in the generative modeling literature, with recent proposals combining transport map learning and importance weight learning in a double robust manner; see \cite{luedtke_doublegen_2025} and references therein. Moreover, in longitudinal treatment settings, interventions can be considered with reference to the \textit{natural value of treatment}, which is the treatment value that would be observed at time $t$ if an intervention were carried out until time $t-1$, but discontinued thereafter. Such values could, in some sense, be viewed as outcomes of the intervention \citep{richardson_single_2013,young_identification_2014,diaz_nonparametric_2023}.
\end{remark}

\subsection{Density ratio and probabilistic classification}
\label{sect:prob}

Here we show how Step 3 of Algorithm \ref{alg:dr} can be recast as an outcome regression problem, enabling implementation using existing software.
Our approach extends existing probabilistic classification methods by \cite{qin_inferences_1998}, \cite{cheng_semiparametric_2004}, \cite{bickel_multi-task_2008}, and \cite{menon_linking_2016}. To motivate a classification-based approach, consider a distribution $Q$ over random variables $(\Delta, X)$ such that $\Delta\in \{0, 1\}$ is a binary random variable, with $P_Q(\Delta=0)=P_Q(\Delta=1)=1/2$, that determines the distribution of $X$, that is $X\sim P_1$ when $\Delta = 1$, and $X\sim P_0$ when $\Delta = 0$. Bayes' Theorem implies that the density ratio is equivalent to the odds
\begin{align}
    \alpha_0(x) &= \frac{P_Q(\Delta=1 \mid X=x)}{P_Q(\Delta=0 \mid X=x)} = \frac{q_0(x)}{1-q_0(x)}. \label{eq:bayes}
\end{align}
where $q_0(x) \equiv P_Q(\Delta=1 \mid X=x)$.
This formulation means that $\alpha_0$ can be learned using probabilistic classification methods that aim to learn $q_0$. This connection is made explicit through Theorem \ref{thrm:prob-to-dens}.

\begin{theorem}
    \label{thrm:prob-to-dens}
    Define the transformed generating function
    \begin{align}
        \tilde{F}(t) \equiv (1-t)F\left(\frac{t}{1-t}\right). \label{eq:f-tilde}
    \end{align}
    Using the shorthand $q(x) \equiv \alpha(x) / \{1+\alpha(x)\}$, we obtain the equality
    \begin{align*}
        \mathcal{R}_F(\alpha)&= 2\E_{Q}\left[ -\tilde{F}\{q(X)\} - \tilde{F}^\prime\{q(X)\}\{\Delta - q(X)\}\right].
    \end{align*}
    See \cite[Lemma 2]{menon_linking_2016} for a related result and Appendix \ref{appendix:transformed_gen_fn} for proof.
\end{theorem}

\begin{corollary}
\label{corollary:prob-to-dens}
The theorem implies that $\mathcal{R}_F(\alpha_0) = 2\E_{Q}\left[-\tilde{F}\{q_0(X)\}\right]$ and the risk difference is $D_F(\alpha_0\|\alpha) = 2\E_{Q}\left[ d_{\tilde{F}}\{q_0(X)\|q(X)\}\right]$.
\end{corollary}

Corollary \ref{corollary:prob-to-dens} shows that the density ratio Bregman divergence implies a classification Bregman divergence that is based on a transformed generating function, see examples in Table \ref{tab:proper-losses}. This is significant because it allows density ratios to be learned in practice using existing algorithms for classification and outcome regression. For example, when $F(t) = t\log(t) - (1+t)\log(1+t)$ corresponds to the negative binomial divergence, Theorem \ref{thrm:prob-to-dens} implies
\begin{align*}
\mathcal{R}_F(\alpha)
&=2\E_{Q}\left[ -(1 - \Delta)\log\{1-q(X)\} - \Delta \log\{q(X)\} \right].
\end{align*}
which is the binary cross-entropy. In the classification literature, unit losses based on Bregman divergences are called proper losses, see e.g. \cite{gneiting_strictly_2007}, \cite{buja_loss_2005}, and \cite{reid_surrogate_2009}. These have a one-to-one correspondence with density ratio Bregman losses, since, for a given $\tilde{F}$, the implied density ratio Bregman divergence is $F(t)= (1+t)\tilde{F}\{t/(1 + t)\}$.

\begin{table}[htbp]
    \centering
    \caption{The transformation in \eqref{eq:f-tilde} applied to functions $F$ in Table \ref{tab:breg}. Each is of the Beta family form $\tilde{F}^{\prime\prime}(t) = t^{a-1}(1-t)^{b-1}$ for constants $a,b$ considered in the probabilistic classification work of \cite{buja_loss_2005} and \cite{zhao_covariate_2019}.}
    \label{tab:proper-losses}
    \begin{tabular}{lll}
        \hline\hline\rule{0pt}{3ex}
        Divergence & $\tilde{F}(t)$ & $\tilde{F}^{\prime\prime}(t)$\\
        \hline
        Least squares & $t^2(1-t)^{-1}/2$ & $(1-t)^{-3}$\\
        Kullback--Leibler & $t\log(t) -t\log(1-t)-t$ & $t^{-1}(1-t)^{-2}$\\
        Negative binomial & $t\log(t) + (1-t)\log(1-t)$ & $t^{-1}(1-t)^{-1}$ \\
        Itakura--Saito & $(1-t)\log(1-t) -(1-t)\log(t)+t-1$ & $t^{-2}(1-t)^{-1}$ \\
        \hline
    \end{tabular}
\end{table}

In view of Theorem \ref{thrm:prob-to-dens}, Step 3 of Algorithm \ref{alg:dr} can be framed as an outcome regression.
To do so, we define the pseudo weight $\omega_i \equiv (\gamma^0_i + \gamma^1_i) / 2$ and pseudo outcome $\tilde{q}_i \equiv \gamma^1_i / (\gamma^0_i + \gamma^1_i)$. Using these, $\mathcal{R}_{F,n}(\alpha)$ in \eqref{eq:rn_normal} becomes
\begin{align}
    \mathcal{R}_{F,n}(\alpha) &= \frac{2}{n}\sum_{i=1}^n \omega_i\left[ -\tilde{F}\{q(x_i)\} - \tilde{F}^\prime\{q(x_i)\}\{\tilde{q}_i - q(x_i)\}\right], \label{eq:rn_pseudo}
\end{align}
which is of the same form as the outcome regression BRR in \eqref{eq:rn_outcome} with observation weights $\omega_i$.
We call $\omega_i$ a weight because $x_i$ weighted by $\omega_i$ represents a sample from the marginal distribution of $X$ under $Q$, but this weight is \textit{pseudo} in the sense that \eqref{eq:rn_pseudo} holds even when $\{(\gamma^0_i, \gamma^1_i)\}_{i=1}^n$ are not normalized.
Additionally, $\tilde{q}_i$ is an outcome since it behaves like an observation from $\Delta$ in expectation, but is \textit{pseudo} in the sense that $\tilde{q}_i \in [0,1]$ but $\Delta \in \{0, 1\}$. Specifically, for a test function $f$
\begin{align*}
    \frac{1}{n} \sum_{i=1}^n \omega_i f(x_i) &\approx \E_{Q}[f(X)] \\
    \frac{1}{n} \sum_{i=1}^n \omega_i \tilde{q}_i f(x_i) &\approx \E_{Q}[\Delta f(X)].
\end{align*}
The form of the BRR estimator in \eqref{eq:rn_pseudo} can be used in Step 3 of Algorithm \ref{alg:dr} to learn $\alpha$, via $q$, using standard software tools that regress $\{\tilde{q}_i\}_{i=1}^n$ on $\{x_i\}_{i=1}^n$ with weights $\{\omega_i\}_{i=1}^n$. Figure \ref{code_block} illustrates this in R using the \verb|stats::glm| solver and a custom \verb|stats::family| object that is available at \url{github.com/CI-NYC/densityratios}. The custom family essentially sets the variance function to be $1/\tilde{F}^{\prime\prime}(t)$, and converts between $q(x)$ and $\alpha(x)$ in the link function. This code snippet can be adapted to use alternative regression methods, such as \verb|glmnet::glmnet| for lasso/ridge regression and \verb|mgcv::gam| for generalized additive modeling.

\begin{figure}[htbp]
\centering
\caption{Example R implementation of Algorithm \ref{alg:dr} Step 3 for estimating the parameter $\theta \in \mathbb{R}^3$ in the parametric density ratio model $\alpha(x) = \exp\{\theta_0 + \theta_1 x_1 + \theta_2 x_2\}$ by empirical minimization of the Kullback--Leibler risk (Divergence \ref{div:kl}). The user provides vectors of numerator/denominator weights and predictors that correspond to $\gamma_i^1, \gamma_i^0$ and $x_i$. The final two lines show how the fitted model can be used to make density ratio predictions on new data.}
\label{code_block}
\begin{lstlisting}
w <- numerator_weights + denominator_weights
pseudo_outcomes <- numerator_weights / w
pseudo_weights <- w / 2

family <- density_ratio_family("kullback-leibler", link="log")
model <- glm(
    formula = pseudo_outcomes ~ 1 + x1 + x2,
    weights = pseudo_weights,
    family = family
)

linear_predictors <- predict(model, newdata, response = "link")
density_ratio_predictions <- family$density.linkinv(linear_predictors)
\end{lstlisting}
\end{figure}

\section{Numerical experiments}
\label{sect:simulations}

In our numerical experiments we consider estimating the average policy effect (Example \ref{example:ape}) with policy $\pi(w) = 1$, the average shift effect (Example \ref{example:ase}) with shift $\delta = 0.1$, and the average dose response curve (Example \ref{example:adrc}) using the importance weighted estimator
\begin{align*}
\frac{1}{n_{\text{eval}}} \sum_{i=1}^{n_{\text{eval}}} y_i \hat{\alpha}(x_i),
\end{align*}
where $\hat{\alpha}$ is the relevant learned density ratio, based on $n_0 = 2{,}000$ training observations and $n_{\text{eval}} = 10{,}000$ evaluation observations. We use a large independent evaluation sample so that our evaluation metrics provide a close approximation to their population analogues. Our main metrics of interest are the absolute bias in the importance weighted estimator versus an oracle where the density ratio is known (i.e., where $\hat{\alpha}$ is replaced with $\alpha_0$ above) and the mean absolute error (MAE) and the root mean squared error (RMSE) in the learned density ratio.
\begin{remark}
By the triangle inequality and Hölder's inequality, the bias in the importance weighted estimator, the MAE, and the RMSE are related as
\begin{align*}
\widehat{\text{Bias}} &\equiv \frac{1}{n_{\text{eval}}} \sum_{i=1}^{n_{\text{eval}}} y_i \{\hat{\alpha}(x_i) - \alpha_0(x_i)\} \\
\lvert\widehat{\text{Bias}}\rvert &\leq \|y\|_{\infty, \text{eval}} \underbrace{ \| \hat{\alpha}(X) - \alpha_0(X) \|_{1, \text{eval}} }_{\text{MAE}} \\
\lvert\widehat{\text{Bias}}\rvert &\leq \|y\|_{2, \text{eval}} \underbrace{ \| \hat{\alpha}(X) - \alpha_0(X) \|_{2, \text{eval}} }_{\text{RMSE}}
\end{align*}
where the empirical p-norm is defined as
\begin{align*}
\|\cdot\|_{p, \text{eval}} \equiv \left(\frac{1}{n_{\text{eval}}} \sum_{i=1}^{n_{\text{eval}}} \lvert (\cdot)_i\rvert^p \right)^{1/p}
\end{align*}
and $p\to \infty$ recovers the supremum norm.
These bounds are insightful, since the empirical bias in the importance weighted estimator depends on the specific outcome generating process $Y \sim P_{Y\mid X}$ in our simulation setup, but the MAE and RMSE do not. Therefore, the MAE and RMSE can also be used to bound other outcome generating processes that we did not consider.
\end{remark}

We compare the following density ratio learners, described in more detail below: a Nadaraya--Watson kernel (NWK) estimator; the uLSIF and KLIEP algorithms; multilayer perceptron (MLP) neural networks and gradient boosted machines (GBMs) for the log density ratio, each trained using one of the divergences from Section \ref{sect:breg}. Additionally, for the binary treatment experiments, we consider  the standard approach of directly learning the propensity score using a probabilistic classifier based on the binary cross-entropy loss. These propensity score methods are denoted MLP-PS and GBM-PS because they use the same MLP and GBM algorithms as the Bregman--Riesz regression methods. Full implementation details are provided in Appendix \ref{sect:implementation_details}.

The NWK estimator is based on the kernel density estimator
\begin{align*}
\hat{p}_X(x) = \frac{1}{n_0} \sum_{i=1}^{n_0} K_X(x, x_i)
\end{align*}
where $K_X(\cdot, \cdot )$ denotes a Gaussian kernel with dimensions of $X$. Using this kernel density estimator, we obtain the NWK estimators
\begin{align*}
\hat{\alpha}^{\text{(ASE)}}(x) = \frac{\hat{p}_X(a - \delta, w)}{\hat{p}_X(a, w)} \\
\hat{\alpha}^{\text{(SW)}}(x) = \frac{\hat{p}_A(a)\hat{p}_W(w)}{\hat{p}_X(a, w)}.
\end{align*}
The uLSIF and KLIEP algorithms yield sieve-based estimators of the form
\begin{align*}
\hat{\alpha}(x) = \sum_{i=1}^p \hat{\theta}_i K_X(x, x_i)
\end{align*}
where $\{x_i\}_{i=1}^p$ represents $p$ random kernel centers, sampled from the numerator observations, and $\hat{\theta}_i \geq 0$ are estimated coefficients. The KLIEP algorithm estimates coefficients by minimizing the Kullback--Leibler BRR under a normality constraint using gradient descent with feasibility satisfaction. The uLSIF algorithm first minimizes the least-squares BRR without constraints, then clips and rescales the resulting coefficients to satisfy the constraint that $\hat{\theta}_i \geq 0$.

The MLP and GBM algorithms learn the log density ratio $\hat{f}$, which implies the density ratio $\hat{\alpha}(x) = \exp\{\hat{f}(x)\}$. For the MLPs, $\hat{f}(x) = f(x\mid \hat{\theta})$ is the output of an MLP neural network with weights and biases $\hat{\theta}$ learned by stochastic gradient descent. For the GBMs, $\hat{f}(x) = \eta \hat{f}_1(x) + \cdots + \eta \hat{f}_K(x)$ represents a sum of $K$ decision trees $\hat{f}_i(x)$ with constant learning rate $\eta$. For both MLP and GBM algorithms, we use early stopping where the $n_0$ observations are split into an 80:20 train-to-validation ratio, with each data split augmented separately. The validation set is also used to tune additional hyperparameters. Both algorithms were trained to optimize an empirical BRR from Section \ref{sect:breg}, which we abbreviate as: least squares (-LS), Kullback--Leibler (-KL), negative binomial (-NB), and Itakura--Saito (-IS).

We consider a data generating process with covariates $X = (A, W)$, where $W \in \mathbb{R}^{20}$ is a vector of predictors with each component $W_i \sim \n(0, 1)$ being independent unit normal and outcome $Y \sim \n(A + AW_1 + W_1 W_2 + W_3, 1)$.
For binary treatment experiments, we let $A$ be Bernoulli distributed with mean $p_{A\mid W}(1\mid w) = \sigma\{|w_1| +(1 - 0.5 w_2)w_3\}$ where $\sigma(u) = 1 / (1 + \exp(-u))$ is the sigmoid function, and for continuous treatment experiments, we let $A \sim \n(c W_1, 1)$, with $c=0.5$. For this data generating process, the average policy effect is $1$,
the average shift effect is $\delta + c = 0.6$,
the average dose response curve is $0$, and the density ratios are $\alpha_0^{\text{(APE)}} = a / p_{A\mid W}(1\mid w)$,
\begin{align*}
\alpha_0^{\text{(ASE)}}(a, w) &= \exp\left(  \delta (a - c w_1) - \frac{\delta^2}{2} \right), \\
\alpha_0^{\text{(SW)}}(a, w) &=  \exp\left(\frac{(a-c w_1)^2}{2} - \frac{a^2}{2(1 + c^2)} -\frac{\log(1 + c^2)}{2}\right).
\end{align*}
The range between the 2.5 and 97.5 percentiles is $(0.0, 2.84)$ for $\alpha_0^{\text{(APE)}}(X)$, $(0.82, 1.21)$ for $\alpha_0^{\text{(ASE)}}(X)$, and $(0.34, 2.37)$ for $\alpha_0^{\text{(SW)}}(X)$.

\subsection{Results}

For the average policy effect, we consider the augmentation scheme described in Augmentation Example \ref{augment:btp}, which we compare against the standard approach where the propensity score is learned directly (MLP-PS and GBM-PS).
The results in Table \ref{tab:results_policy} show that density ratio learning using Bregman--Riesz regression appears to perform better than propensity score-based methods in terms of our primary metrics. Additionally, we see that the choice of the Bregman divergence has some impact on learning performance, with the least squares divergence generally performing worse than the other divergences. This phenomenon is possibly explained by appealing to the weight expressions $F^{\prime\prime}$ in Table \ref{tab:breg}, with the least squares divergence assigning a constant weight to all values of the density ratio, versus the other divergences where the weight decreases as the density ratio increases. Consequently, it is possible that the least-squares divergence is overfitting to large density ratio values, degrading overall performance. Conversely, the negative binomial and Itakura--Saito divergences, which give relatively less weight to large density ratio values, appear to perform better for both the MLP and GBM learners.

For the average shift effect, we consider the modified treatment policy augmentation scheme presented in Augmentation Example \ref{augment:mtp}. The results in Table \ref{tab:results_shift}, show that the KDE gives the best performance in terms of absolute bias; however, the MLP neural networks outperform all other algorithms in terms of MAE and RMSE, suggesting that MLP learners might generalize better to other outcome distributions not considered in this study. The choice of Bregman divergence appears to have no clear effect on MLP and GBM performance in this experiment, which could be because the true density ratios $\alpha_0^{\text{(ASE)}}(X)$ are concentrated relatively close to 1, compared with $\alpha_0^{\text{(APE)}}(X)$ and $\alpha_0^{\text{(SW)}}(X)$, and hence overfitting to larger density ratio values is less of a concern.

For the average dose response curve, we compare sampling with replacement ($n_1 = m n_0$) versus m-permutations and m-derangements for $m \in \{1, 2, 5, 10\}$ as described in Augmentation Example \ref{augment:sw}. For the MLP, we also compare train-time sampling (MLP-TT) with replacement/permutation/derangement, each with $n_1 = n_0$. For the MLP-TT results, the multiplier $m$ affects only the sampling scheme in the validation split of the training data.

\begin{figure}[htbp]
    \centering
    \includegraphics[width=1.0\linewidth]{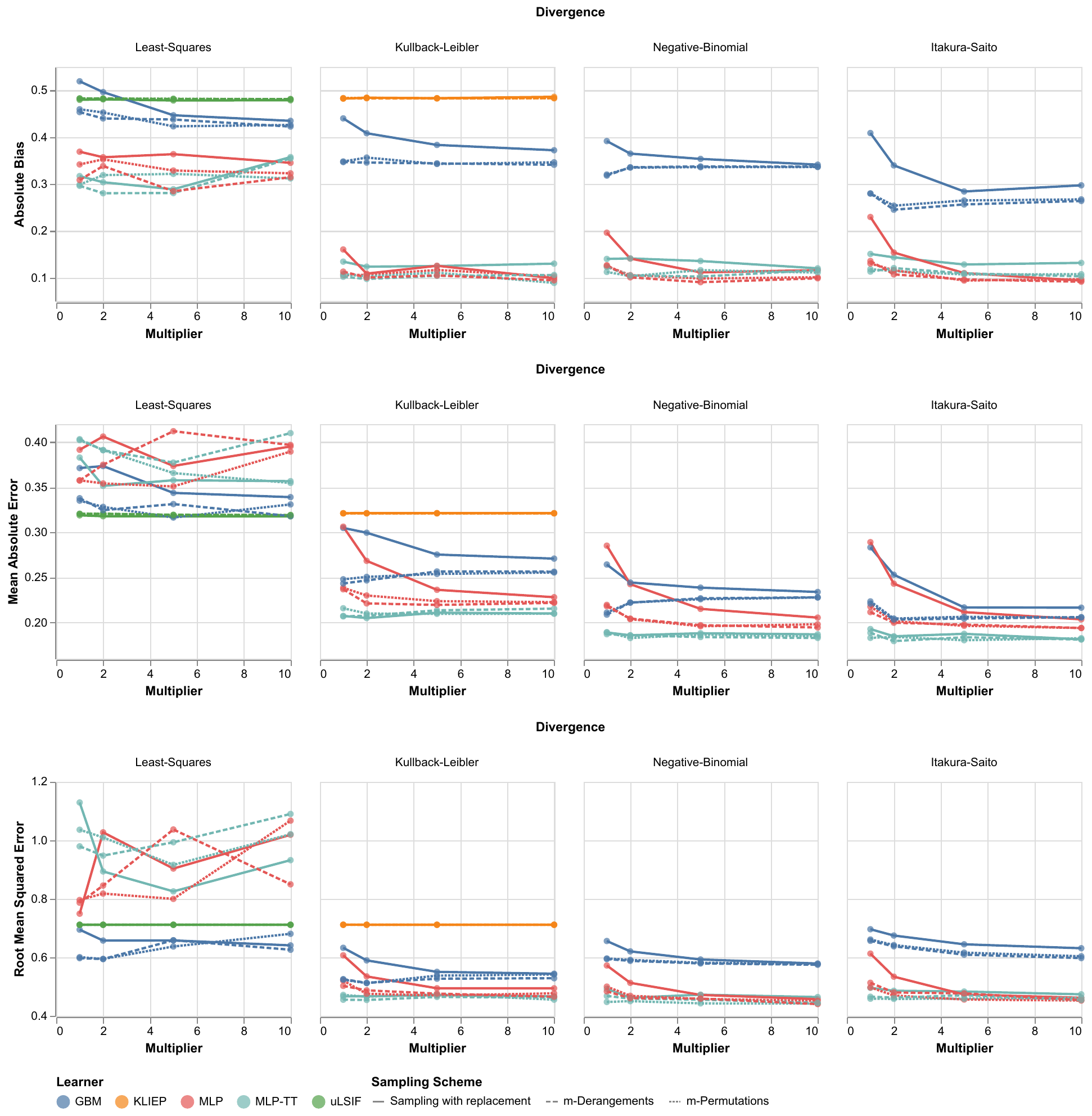}
    \caption{Absolute bias of the importance weighted average dose response curve estimator and MAE and RMSE in the corresponding stabilized weights versus multiplier $m$ for each learner and sampling scheme. Each point represents a median over 100 dataset replicates. Results for the KDE, which do not depend on the sampling scheme, are not shown because they far exceed the axis limits in the MAE and RMSE plots (absolute bias: 0.390, MAE: 1.35, RMSE: 16.3).}
    \label{fig:sw}
\end{figure}

The results in Figure \ref{fig:sw} show that the least squares divergence generally performs poorly for stabilized weight learning, possibly due to the least squares divergence overfitting to large density ratio values, as described above. Due to the noisy performance of the least squares divergence, we consider only the negative binomial, Itakura--Saito, and Kullback--Leibler divergences when drawing conclusions about the effect of the sampling scheme on learning stabilized weights. We see that the m-permutation and m-derangement methods give broadly similar results for all learners, and that for the MLP and GBM learners, m-permutation and m-derangement both tend to outperform the sampling with replacement method, though this improvement attenuates as the multiplier $m$ increases. The MLP-TT learners have the best overall performance in this study, with no clear difference between m-permutation, m-derangement, and sampling with replacement. Moreover, the multiplier $m$ has limited effect on the MLP-TT learners, which is to be expected since $m$ affects only the validation split and not the training split for the MLP-TT algorithms.

Based on this numerical study we recommend the negative-binomial or Itakura--Saito divergences in settings where overfitting to large density ratio values may be a concern.
One expects large density ratio values when there is poor overlap between the numerator and denominator distributions---a situation where Riesz representer and density ratio learning is known to be more challenging \citep{damour_overlap_2021}.
For learning stabilized weights, we conclude that train-time augmentation with even a small validation set ($m=1$) gives improved performance over using fixed sampling schemes. When train-time augmentation is not possible, however, we recommend m-permutations or m-derangements sampling with a modest multiplier $(m \geq 2)$.

\begin{table}[htbp]
\centering
\caption{Results for estimation of the average policy effect ($\pi(w) = 1$) and corresponding density ratio estimates obtained as the median over $100$ dataset replicates. Learners are sorted by absolute bias (AB) in the importance weighted estimator.}
\label{tab:results_policy}
\begin{tabular}{llll}
\hline\hline
Learner & AB & MAE & RMSE \\
\hline
% KDE & 2.32 & 4.10 & 107.0 \\
MLP-PS & 0.822 & 0.671 & 0.983 \\
GBM-PS & 0.63 & 0.585 & 0.901 \\
uLSIF & 0.269 & 0.276 & 0.705 \\
KLIEP & 0.249 & 0.267 & 0.703 \\
GBM-LS & 0.244 & 0.265 & 0.617 \\
GBM-KL & 0.188 & 0.224 & 0.585 \\
GBM-NB & 0.121 & 0.171 & \textbf{0.539} \\
MLP-LS & 0.0778 & 0.257 & 0.674 \\
GBM-IS & 0.0764 & \textbf{0.158} & 0.568 \\
MLP-KL & 0.0448 & 0.221 & 0.599 \\
MLP-NB & 0.0387 & 0.209 & 0.592 \\
MLP-IS & \textbf{0.0343} & 0.207 & 0.578 \\
\hline
\end{tabular}
\end{table}

\begin{table}[htbp]
\centering
\caption{Results for estimation of the average shift effect ($\delta = 0.1$) and corresponding density ratio estimates, obtained as the median over $100$ dataset replicates. Learners are sorted by absolute bias (AB) in the importance weighted estimator.}
\label{tab:results_shift}
\begin{tabular}{llll}
\hline\hline
Learner & AB & MAE & RMSE \\
\hline
uLSIF & 0.0955 & 0.0798 & 0.0997 \\
GBM-IS & 0.0933 & 0.0773 & 0.0978 \\
KLIEP & 0.0910 & 0.0738 & 0.0933 \\
GBM-KL & 0.0845 & 0.0778 & 0.0985 \\
GBM-NB & 0.0826 & 0.0772 & 0.0978 \\
GBM-LS & 0.0619 & 0.0766 &  0.0969 \\
MLP-IS & 0.0151 & 0.0381 & 0.0483 \\
MLP-KL & 0.0146 & 0.0379 & 0.0480 \\
MLP-NB & 0.0135 & \textbf{0.0376} & \textbf{0.0480} \\
MLP-LS & 0.0130 & 0.0383 & 0.0488 \\
KDE & \textbf{0.0113} & 0.0596 & 0.0782 \\
\hline
\end{tabular}
\end{table}

\pagebreak
\section{Related work}
\label{sect:related}

\textbf{Balancing weights}: A popular approach for approximating Riesz representers in causal inference is to learn a \textit{balancing weight} function $\tilde{\alpha}$ such that $\langle f, \tilde{\alpha} \rangle \approx \h(f)$ for test functions $f \in \mathcal{H}$. Following \cite{wang_minimal_2019} and \cite{ben-michael_balancing_2021} several such balancing weights can be characterized, in the population limit, as
\begin{align*}
    \tilde{\alpha} \equiv \argmin_{\alpha \in \mathcal{H}} E_{P_0}[F\{\alpha(X)\}] + \Lambda(\alpha_0 - \alpha)
\end{align*}
where $\Lambda: \mathcal{H} \to \mathbb{R} \cup \{\infty\}$ is convex with a minimum at $\Lambda(0) = 0$. Examples for $F$ include entropy balancing \citep{hainmueller_entropy_2012} with $F(t) = t\log(t) - t$, and  stable balancing \citep{Zubizarreta2015} with $F(t) = (t - 1)^2$. For $\Lambda$, consider the norm constraint example, where for a norm $\|\cdot\|$ on $\mathcal{H}$, and constant $\lambda >  0$, $\Lambda(h) = 0$ if $\|h\|\leq \lambda$ and $\Lambda(h) = \infty$ otherwise.

Several duality results have been shown for balancing weights, often appealing to Lagrangian duality when $\Lambda$ represents a constraint \citep{zhao_covariate_2019,wang_minimal_2019}. Following the Fenchel duality approach of \cite{ben-michael_balancing_2021}, balancing weights can be expressed via the dual problem
\begin{align}
    \tilde{\alpha} = \argmin_{\alpha \in \mathcal{H}} \mathcal{R}_{F}(\alpha) + \Lambda^*\left(F^\prime \circ \alpha \right), \label{eq:dual_bw_main}
\end{align}
where $\Lambda^*$ is the convex conjugate of $\Lambda$, see Appendix \ref{sect:duality} for details. For example, letting $\|\cdot\|^*$ denote the dual norm of $\|\cdot\|$, then the norm constraint corresponds with $\Lambda^*(h) = \lambda \|h\|^*$. The parameter $\lambda$ can be viewed as a regularization parameter, controlling the contribution of the regularizer $\|F^\prime \circ \alpha\|^*$, with the unregularized Bregman--Riesz regression problem recovered in the limit $\lambda \to 0$. We therefore interpret balancing weight learning as a special case of penalized Bregman--Riesz regression, with a regularizer $\Lambda^*\left( F^\prime \circ \alpha\right)$ that acts on the link function scale, see Remark \ref{remark:duality}.

\textbf{Pseudo-outcome learners}: The Bregman--Riesz regression framework can be used to understand pseudo-outcome learners of the conditional average treatment effect and the dose-response curve due to \cite{kennedy_towards_2023} and \cite{Kennedy2017}.
For $X=W$, the conditional average treatment effect $\tau(w) \equiv \E(Y^1-Y^0 \mid W=w)$ is the Riesz representer of the map $\h(f) = \E[f(W)(Y^1-Y^0)]$, which is a bounded linear functional when $\tau \in \mathcal{H}$.
Similarly, for $X=A$, the dose-response curve $\psi(a) \equiv \E[Y^a]$ is the Riesz representer of $\h(f) = \E[f(A)\psi(A)]$, assuming that $\psi \in \mathcal{H}$. In both cases, Bregman--Riesz regression learners can be constructed using an estimator of $\h(f)$. The two pseudo-outcome learners by Kennedy et al. are of this form, where $\h$ is estimated using a double robust estimator with cross-fitted nuisance functions, and additive constants in the BRR are ignored. The so called EP-learner of the conditional average treatment effect, due to \cite{van_der_laan_combining_2024}, is also of this form, but uses targeted maximum likelihood estimation (TMLE) to estimate $\h$. Specifically, the TMLE estimator targets $\h(f_j)$ for a set of basis functions indexed by $j$, resulting in an estimator of $\h(f)$ that is efficient and double robust when $f$ is a linear combination of the basis function.

\textbf{Time Score Matching}: One of the fundamental challenges in density ratio learning is dealing with poor overlap between the numerator and denominator distributions, known as the \textit{density chasm} problem. To overcome this, \cite{rhodes_telescoping_2020} propose a telescopic density ratio estimator, which decomposes the density ratio into a product of simpler interpolating density ratios. This approach is extended in the infinitesimal limit by \cite{choi_density_2022}, who characterize the log density ratio as the integral
\begin{align*}
\log\{\alpha_0(x)\} = \log\left\{\frac{p_1(x)}{p_0(x)}\right\}  = \int_0^1 \frac{\partial}{\partial t} \log\{p_t(x)\} \dd t
\end{align*}
where $p_t(x)$ is a user-defined path between $p_0(x)$ and $p_1(x)$ with univariate parameter $t \in [0, 1]$. Rather than learn the density ratio, they propose learning the \textit{time score} function $\frac{\partial}{\partial t} \log\{p_t(x)\}$ and then using numerical integration to approximate the log density ratio. Variants of this method, with additional conditioning and path proposals, are studied by \cite{yu_density_2025}, with further connections to flow matching and optimal transport techniques, see references therein.

\textbf{Other learning algorithms}: In the current work we provide an overview of several foundational density ratio learning approaches; however, there is a rich literature of recent methods which we are unable to cover in detail. In particular we highlight the following papers for further study.
\cite{sugiyama_direct_2011} propose learning the density ratio in a linear subspace of $\mathcal{X}$ via dimensional reduction.
\cite{izbicki_high-dimensional_2014} propose a spectral algorithm that expands the density ratio in terms of the eigenfunctions of a kernel-based operator.
\cite{choi_featurized_2021} use a generative neural network model to learn the density ratio in a latent space where the numerator and denominator distributions have improved overlap.
\cite{nguyen_regularized_2024} extend the regularization theory for density ratio estimators in reproducing kernel Hilbert spaces.
\cite{wang_projection_2025} propose learning density ratios using a modified projection pursuit algorithm. \cite{huk_your_2025} describe connections between copula-based density ratio models and probabilistic classification. In our previous work in the average linear functional setting we consider a constraint-based interpretation of Riesz regression \citep{hines_automatic_2025}.

\textbf{Calibration}: In addition to Riesz regression, recent work has proposed using the Riesz regression loss to calibrate Riesz representer estimates in order to reduce bias in estimators that rely on cross-fitted Riesz representer estimates \citep{van_der_laan_doubly_2025,van_der_laan_generalized_2025}. For instance, consider that the data sample is split into a Riesz representer training sample and an evaluation sample. The uncalibrated procedure learns $\hat{\alpha}$ in the training sample, then evaluates the estimator in the evaluation sample using $\hat{\alpha}$. The calibrated procedure still learns $\hat{\alpha}$ in the training sample, but instead evaluates the estimator in the evaluation sample using $\hat{\alpha}^* = \hat{f} \circ \hat{\alpha}$, where $\hat{f}: \mathbb{R} \to \mathbb{R}$ is an isotonic (monotonic) calibration function that is learned in the evaluation sample by Riesz regression. In principle, different Riesz losses could be considered for the calibration step, such as those described in the current paper; however, more work is required to verify that the theoretical benefits of calibration are maintained.

\section{Discussion}
\label{sect:discussion}

Through our numerical experiments, we find that the negative binomial, Kullback--Leibler, and Itakura--Saito divergences tend to outperform the least-squares divergence when learning density ratios that take large values, as is common in causal problems where overlap between the numerator and denominator distributions is poor.
We attribute this finding to the weight $F^{\prime\prime}(t)$ in \eqref{eq:bman_cost}, which is constant for the least-squares divergence but decreasing in $t$ for the other divergences.
Our discussion of data augmentation clarifies the need for augmentation observed by \cite{diaz_nonparametric_2023} and \cite{Lee2025} when learning Riesz representers, and demonstrates how augmentation can be achieved for several causal problems in practice.
For the stabilized weight density ratio, we find that train-time sampling outperforms fixed sampling schemes, and that m-permutations or m-derangements with $m \geq 2$ are preferred over sampling with replacement.
Through Theorem \ref{thrm:prob-to-dens} we formally connect density ratio learning with an implied classification problem, enabling implementation using standard outcome regression libraries.

Several directions remain open for future work.
We cover augmentation strategies for canonical examples in causal inference, but strategies for more complex interventions, e.g. in longitudinal settings, may require bespoke schemes.
The weight $F^{\prime\prime}$ provides a clear handle for adapting the Bregman--Riesz risk to the structure of $\alpha_0$, suggesting an iterative procedure in which an initial density ratio learner is used to inform the choice of $F$ for a refined fit.
Analogous procedures for linear regression have recently been proposed by \cite{feng_optimal_2025}, where the distribution of regression residuals is used to inform a subsequent regression loss.
A separate question concerns the extension of Bregman--Riesz regression to signed Riesz representers, such as that of the average treatment effect, which require generating functions that are convex on the full real line. Remark \ref{remark:ate} shows how the Itakura--Saito divergence can be modified to learn signed Riesz representers when the sign is structurally known; the general case, where the sign of $\alpha_0$ is unknown, remains open.
Within causal inference, the duality between Bregman--Riesz regression and balancing weight learning described in Section \ref{sect:related} suggests that loss-based and balancing-based approaches can be combined by using the regularizer $\Lambda^*(F^\prime \circ \alpha)$ to target test functions of practical interest.
Moreover, several density ratio learning methods discussed in Section \ref{sect:related} are not directly captured by the Bregman--Riesz framework, and their integration represents an attractive direction for future work.
Finally, the consistency and rate guarantees that justify the use of density ratio learners in downstream debiased estimation \citep{Hirshberg2021,chernozhukov_automatic_2024} are inherited from existing theory rather than extended here, with sharp rates that exploit the specific choice of Bregman divergence remaining an open theoretical question.

\section*{Acknowledgments}

We thank Ivan Díaz, Alejandro Schuler, Avi Feller and Eli Ben-Michael for helpful conversations and correspondence on this topic.

\bibliographystyle{apalike}
\bibliography{refs}

\pagebreak
\appendix
\setcounter{equation}{0}
\renewcommand{\theequation}{\thesection.\arabic{equation}}
\renewcommand{\theHequation}{\thesection.\arabic{equation}}
\renewcommand{\theHfigure}{\thesection.\arabic{figure}}
\renewcommand{\theHtable}{\thesection.\arabic{table}}

\section{Numerical study implementation details}
\label{sect:implementation_details}

Here we outline implementation details for the numerical study presented in Section \ref{sect:simulations}. Full replication code is available at \url{https://github.com/CI-NYC/densityratios}.

For Kernel Density Estimation we use the \verb|jax.scipy.stats.gaussian_kde| function available in version \verb|0.7.2| of JAX (\url{http://github.com/google/jax}). To determine the bandwidth we use the default `scott' rule of thumb.

For the KLIEP and uLSIF algorithms we re-implement the reference implementations from \url{https://github.com/srome/pykliep} and \url{https://github.com/hoxo-m/densratio_py} respectively. Our implementation adds a convergence criteria to the reference KLIEP algorithm (coefficients change by less than $10^{-6}$ in consecutive iterations), and use JAX function compilation to reduce runtime. The KLIEP and uLSIF implementations use the hyperparameter search algorithms described in the original papers, respectively: \citet[Figure 1]{sugiyama_direct_2007} and \citet[Figure 2]{kanamori_least-squares_2009}. We search over the hyperparameter values described in Table \ref{tab:hyperhyper}.

For the GBM learners we use version \verb|4.6.0| of LightGBM (\url{https://github.com/microsoft/LightGBM}). The number of trees (maximum 1000) is determined by early stopping, via loss stagnation in a validation set (minimum improvement of 0, patience 10). The validation set is also used to tune hyperparameters according to the search grid described in Table \ref{tab:hyperhyper}. Specifically, the LightGBM training algorithm is run for each hyperparameter combination and the learner with the lowest validation loss is selected.

For the MLP learners we use a custom implementation, built using version \verb|2.8.0| of Pytorch (\url{https://github.com/pytorch/pytorch}). MLP weights and biases are trained via stochastic gradient descent (batch size 125, ADAM optimizer, learning rate $10^{-4}$). The number of training epochs (maximum 1000) is determined by early stopping, via loss stagnation in a validation set (minimum improvement of 0, patience 5). The validation set is also used to tune hyperparameters according to the search grid described in Table \ref{tab:hyperhyper}. Specifically, our training algorithm is run for each hyperparameter combination and the learner with the lowest validation loss is selected.

\begin{table}[htbp]
\caption{Hyperparameter search values for each learning algorithm.}
\label{tab:hyperhyper}
\centering
\begin{tabular}{lll}
\hline\hline
Learner & Hyperparameter & Grid \\
\hline
KLIEP/uLSIF & basis dimension & $\{100, 200, 500\}$ \\
KLIEP/uLSIF & bandwidth & $\{1, 2, 5, 10\}$ \\
uLSIF & smoothing parameter & $\{0.0, 0.1, 0.5, 1.0\}$ \\
GBM & maximum tree depth & $\{10, 20, 50\}$ \\
GBM & learning rate & $\{10^{-3}, 10^{-4}\}$ \\
GBM & bagging fraction & $\{0.8, 1.0\}$ \\
MLP & depth & $\{2, 3\}$ \\
MLP & width & $\{20, 50\}$ \\
\hline
\end{tabular}
\end{table}

\section{Score matching}
\label{sect:score_matching}

Here we show how the Riesz representer setting described in Section \ref{sect:setup} generalizes when the Hilbert space $\mathcal{H}$ is made up of vector functions. Such an extension forms the basis of \textit{score matching} techniques used in score-based generative diffusion models \citep{hyvarinen_estimation_2005,lyu_interpretation_2009,song_generative_2019}.

The setup extends as follows: consider a random variable $X \in \mathcal{X}$ distributed according to an unknown distribution $P_0$. Let $\mathcal{H}$ be a Hilbert space of real-valued functions $f:\mathcal{X} \to \mathbb{R}^d$ with inner product $\langle f, g\rangle \equiv \E_{P_0}[f(X)^\top g(X)]$ for $f, g \in \mathcal{H}$. Letting $\h: \mathcal{H} \to \mathbb{R}$ be a continuous and linear map, Riesz representation theorem implies that there exists a unique Riesz representer $\alpha_0 \in \mathcal{H}$ such that $\h(f) = \langle f, \alpha_0 \rangle$ for all $f\in \mathcal{H}$. Moreover, the Bregman divergence in \eqref{defn:breg} generalizes as
\begin{align*}
D_F(\alpha_0 \| \alpha) = \E_{P_0}\left[F\{\alpha_0(X)\} - F\{\alpha(X)\} - \nabla F \{\alpha(X)\}^\top \{\alpha_0(X) - \alpha(X)\}\right]
\end{align*}
where $F: \mathbb{R}^d \to \mathbb{R}$ is a convex differentiable function with Jacobian $\nabla F$. Discarding the constant term $\E_{P_0}\left[F\{\alpha_0(X)\}\right]$, one obtains the BRR
\begin{align*}
\mathcal{R}_F(\alpha) &= \E_{P_0}\left[\nabla F\{\alpha(X)\}^\top \alpha(X) - F\{\alpha(X)\}\right] - \h(\nabla F \circ \alpha).
\end{align*}
For score matching, consider the setting where $X$ is a continuous random variable over $\mathcal{X} \subseteq \mathbb{R}^d$, and we define the linear map $\h(f) = -\E_{P_0}[\Tr\{\nabla f(X)\}]$ where $\nabla f$ is the $d\times d$ Jacobian matrix of $f$ and $\Tr$ denotes the matrix trace. Note that this linear map is of the average linear functional type described in Setting \ref{setting:alf} of Section \ref{sect:setup}, and that the trace of the Jacobian is also called the \textit{divergence} in vector calculus. Under mild regularity conditions, including that $p_X$ is differentiable, one can write $\h(f) = \langle f, \alpha_0^{\text{(SM)}} \rangle$ where $\alpha_0^{\text{(SM)}}(x) = \nabla \log\{p_X(x)\}$.

When $F(t) = \|t\|^2$ is the squared Euclidean distance, with $\nabla F(t) = 2 t$, then one obtains the least-squares divergence and BRR
\begin{align*}
D_F(\alpha_0 \| \alpha) &= \E_{P_0}\left[ \|\alpha(X) - \alpha_0(X)\|^2\right] \\
\mathcal{R}_F(\alpha) &= \E_{P_0}\left[ \|\alpha(X)\|^2 +2 \Tr\{\nabla \alpha(X) \}\right].
\end{align*}
The score matching procedure minimizes an empirical approximation of this least-squares BRR.

\section{Properties of smooth Bregman divergences}
\label{sect:breg_proof}

Here we prove two standard statements about Bregman divergences in the case where $F$ is twice differentiable, with second derivative $F^{\prime\prime}$. The first relates to the integral representation of the Bregman divergence in the main text, and the second is a strong convexity result.

\textbf{Claim}:
\begin{align*}
d_F(v_0\|v_1) &= \int_{-\infty}^\infty r(t\mid v_0, v_1) F^{\prime\prime}(t)\dd t
\end{align*}
\begin{proof}
The derivative of $(t - v_0)F^\prime(t) - F(t)$ with respect to $t$ is $(t-v_0) F^{\prime\prime}(t)$. Hence
\begin{align*}
    d_F(v_0\|v_1) &= F(v_0) - F(v_1) - F^\prime(v_1)(v_0 - v_1) \\
    &= \int^{v_1}_{v_0} (t-v_0) F^{\prime\prime}(t) \dd t \\
    &= \int_{\min(v_0,v_1)}^{\max(v_0, v_1)} |t-v_0| F^{\prime\prime}(t) \dd t \\
    &= \int_{-\infty}^\infty r(t\mid v_0, v_1) F^{\prime\prime}(t) \dd t.
\end{align*}
\end{proof}

\textbf{Claim}: Assume that $\alpha_0(X) \in [a, b]$ and $\alpha(X) \in [a, b]$ with probability 1 for constants $a$ and $b$. Then, for $k_F \equiv 2 \inf_{t \in [a, b]} F^{\prime\prime}(t)$
\begin{align*}
    k_F\|\alpha_0 - \alpha\|_{\mathcal{H}}^2 \leq D_F(\alpha_0 \|\alpha).
\end{align*}
\begin{proof}
From the previous proof, for $v_0, v_1 \in [a, b]$
\begin{align*}
    d_F(v_0\|v_1) &= \int^{v_1}_{v_0} (t-v_0) F^{\prime\prime}(t) \dd t \\
    &\geq k_F \int^{v_1}_{v_0} (t-v_0) \frac{1}{2} \dd t \\
    &= k_F (v_1 - v_0)^2
\end{align*}
Since $D_F(\alpha_0 \|\alpha) = \E[d_F\{\alpha_0(X)\|\alpha(X)\}]$ the result holds.
\end{proof}

\section{Generalized linear models and the Bregman divergence}
\label{sect:glm_deviance}

Here we review Generalized Linear Model (GLM) theory as it relates to the Bregman divergences in the main text. See \citet[Chapter 5]{dunn_generalized_2018} for a GLM reference text, and e.g. \citet[Theorem 3]{banerjee_clustering_2004} for details on the link between Bregman divergences and exponential families.
% and \cite{wedderburn_quasi-likelihood_1974}.

Let $Y \in \mathbb{R}$ be distributed according to a one-parameter exponential family distribution with canonical parameter $\theta_0 \in \Theta \subseteq \mathbb{R}$ and density
\begin{align*}
    p^*(y\mid \theta_0) = a(y, \phi) \exp\left\{ \frac{y\theta_0 - \kappa(\theta_0)}{\phi}\right\}
\end{align*}
where $\kappa: \Theta \to \mathbb{R}$ is a known cumulant function, the dispersion parameter $\phi > 0$ is known, and $a(y,\phi)$ is a function of $y$ and $\phi$ only.
This distribution has the mean and variance $\mu_0 \equiv \E[Y] = \kappa^\prime(\theta_0)$ and $\Var(Y) = \phi \kappa^{\prime\prime}(\theta_0)$. Since $\kappa^{\prime\prime}(\theta) > 0$ is a variance, $k^\prime(\theta)$ is strictly increasing with an inverse, called the link function, which we denote $g$, with $\theta_0 = g(\mu_0)$. The variance function is defined as $V(\mu) = \kappa^{\prime\prime}\left\{g(\mu)\right\}$ with $\Var(Y) = \phi V(\mu_0)$.

The exponential family density can be reparameterized in terms of the convex conjugate of $\kappa$, defined as
\begin{align*}
    F(\mu) &\equiv \sup_{\theta\in\Theta} \mu \theta - \kappa(\theta) \\
    &= \mu g(\mu) - \kappa\{g(\mu)\}
\end{align*}
with $F^\prime(\mu) = g(\mu)$ and $F^{\prime\prime}(\mu) = 1/V(\mu)$. Using this function, the one parameter exponential density can be written in terms of its mean as
\begin{align*}
    p(y\mid \mu_0) = a(y, \phi) \exp\left\{ \frac{F(\mu_0) + F^\prime(\mu_0)(y-\mu_0)}{\phi}\right\}
\end{align*}
Example densities of this form are described in Table \ref{tab:exponential-families}.
For an alternative mean value $\mu$, one can write
\begin{align*}
    p(y\mid \mu_0) &=a(y, \phi) \exp\left\{\frac{F(\mu)}{\phi}\right\} \exp\left\{ - \frac{F(\mu) - F(\mu_0) - F^\prime(\mu_0)(y-\mu_0)}{\phi}\right\}
\end{align*}
Since the left does not depend on $\mu$, this expression must also hold in the limit $\mu \to y$, which is convenient, since it relates the numerator in the second exponential above to the Bregman divergence with generating function F. Specifically,
\begin{align*}
    p(y\mid \mu_0) &= b(y, \phi)\exp\left\{ \frac{-d(y,\mu_0)}{2\phi}\right\}
\end{align*}
where $b(y, \phi) \equiv a(y,\phi) \exp\{F(y)/\phi\}$ and we define the unit deviance
\begin{align*}
    d(y,\mu) &\equiv 2\{F(y) - F(\mu) - F^\prime(\mu)(y - \mu)\} \\
    &= 2 d_F(y \|\mu)
\end{align*}
Given $n$ iid observations of $Y$, one obtains the negative log likelihood
\begin{align*}
    \sum_{i=1}^n -\log\{p(y_i\mid \mu_0)\} &= \frac{1}{2\phi }\sum_{i=1}^nd(y_i, \mu_0) - \sum_{i=1}^n \log\{b(y_i, \phi)\}
\end{align*}
Hence, the maximum likeilhood estimator for $\mu_0$ minimizes to total deviance
\begin{align*}
    \frac{1}{n}\sum_{i=1}^n d(y_i, \mu) = \frac{2}{n}\sum_{i=1}^n d_F(y_i \|\mu)
\end{align*}
over $\mu$. In the generalized linear model framework, this analysis is extended to model $\mu_i = h^{-1}(\eta_i)$ where $h$ is a link function and $\eta_i = \beta^\top x_i$ is a linear predictor with parameter vector $\beta$ and covariates $x_i$ that may include an intercept. The canonical link function $h = g$ is often used but not required.

To conclude, the Bregman convex generating function $F$ implies a weight $F^{\prime\prime}(t)$ for model errors, which is analogous to the variance function $V(\mu)$ for generalized linear model regression. Moreover, $F^\prime$ can be viewed as a canonical link function associated with $F$.

\begin{table}[htbp]
    \centering
    \caption{Examples of the function $F$ used in generalized linear modeling, which are convex and twice differentiable on the given support. Each $F$ implies a variance function $V(\mu) = 1/F^{\prime\prime}(\mu)$ and a canonical inverse link function $g^{-1}$, defined as the inverse of $F^\prime$.
    Each function is also associated with a one parameter exponential family distribution. Examples where $V(\mu) = \mu^a$, for fixed value $a$, are members of the Tweedie distribution family \citep{jorgensen_exponential_1987}. For the negative binomial distribution, the `number of failures' parameter is fixed to 1.}
    \label{tab:exponential-families}
    \begin{tabular}{lllll}
    \hline\hline
        Family & Support & $F(\mu)$ & $V(\mu)$ & Canonical $g^{-1}(\eta)$ \\ \hline
        Gaussian & $\mathbb{R}$ & $\mu^2/2$ & $1$ & $\eta$ \\
        Poisson & $\mathbb{R}_{>0}$ & $\mu\log(\mu) - \mu$ & $\mu$ & $\exp(\eta)$ \\
        Gamma & $\mathbb{R}_{>0}$ & $-\log(\mu) - 1$ & $\mu^2$ & $-\eta^{-1}$ \\
        Inverse Gaussian & $\mathbb{R}_{>0}$ & $(2\mu)^{-1}$ & $\mu^3$ & $(-2\eta)^{-1/2}$ \\
        Binomial & $(0,1)$ & $\mu\log(\mu) + (1-\mu)\log(1-\mu)$ & $\mu(1-\mu)$ & $\{\exp(-\eta)+1\}^{-1}$ \\
        Negative Binomial & $\mathbb{R}_{>0}$ & $\mu\log(\mu) - (1+\mu)\log(1+\mu)$ & $\mu(1+\mu)$ & $\{\exp(-\eta) - 1\}^{-1}$ \\
    \hline
    \end{tabular}
\end{table}

\section{Symmetry of Bregman divergences}
\label{sect:sym}

Here we discuss the symmetry of Bregman divergences when the role of $P_0$ and $P_1$ are reversed. Provided that $P_0 \ll P_1$, the resulting density ratio is
\begin{align*}
    \frac{p_0(x)}{p_1(x)} &= \frac{1}{\alpha_0(x)}.
\end{align*}
Swapping $P_0$ and $P_1$ in the Bregman divergence requires taking expectations of the unit Bregman divergence over $P_1$. That is, for a pair of function $g_0, g \in \mathcal{H}$, we define
\begin{align*}
    D^{\text{(rec)}}_F\left(g_0 \| g \right) &\equiv \E_{P_1}\left[d_F\{g_0(X)\|g(X)\}\right].
\end{align*}
where `rec' stands for reciprocal. Letting $1/\alpha$ denote the function $x \mapsto 1/\alpha(x)$, we find that
\begin{align*}
D^{\text{(rec)}}_F\left(\frac{1}{\alpha_0} \middle\| \frac{1}{\alpha}\right) = D_{F^{\text{(rec)}}}\left(\alpha_0 \| \alpha\right)
\end{align*}
where $F^{\text{(rec)}}(t) \equiv t F(1/t)$, which is convex for $t > 0$ when $F(t)$ is itself convex for $t > 0$.
In other words: swapping the roles of $P_0$ and $P_1$ is equivalent to choosing a different convex generating function $F^{\text{(rec)}}$.
For example, for $F(t) = -\log(t) -1$ one obtains $F^{\text{(rec)}}(t) = t \log(t) - t$. Therefore, the Itakura--Saito distance is the same as the Kullback--Leibler divergence when the roles of $P_0$ and $P_1$ are reversed.
For $F(t) = t \log(t) - (1+t) \log(1+t)$, one obtains  $F^{\text{(rec)}}(t) = F(t)$ hence the negative binomial divergence is invariant to reversing the roles of $P_0$ and $P_1$. Table \ref{tab:sym_F} shows the form of $F^{\text{(rec)}}$ for several examples.

\begin{table}[htbp]
\caption{Reciprocal convex generator functions $F^{\text{(rec)}}$ for the divergences in the main text. The name of the Reciprocal divergence is also given, with the Inverse Gaussian name following the convention in Table \ref{tab:exponential-families}.}
\label{tab:sym_F}
\centering
\begin{tabular}{llll}
\hline\hline
Divergence & Reciprocal divergence & $F(t)$ & $F^{\text{(rec)}}(t)$ \\ \hline
Least squares & Inverse Gaussian & $t^2/2$ &  $(2t)^{-1}$ \\
Kullback--Leibler & Itakura--Saito & $t \log(t) - t$ &  $-\log(t) - 1$ \\
Negative binomial & Negative binomial & $t \log(t) - (1+t) \log(1+t)$ &  $F(t)$ \\
Itakura--Saito & Kullback--Leibler & $-\log(t) - 1$ &  $t\log(t) - t$ \\
\hline
\end{tabular}
\end{table}

\section{Mapping density ratios to the probability scale}
\label{appendix:transformed_gen_fn}

Here we prove a general version of Theorem \ref{thrm:prob-to-dens}, which shows why the factor of 2 appears in the original statement. Specifically, we let $P_Q(\Delta = 1) = \rho$ be a constant on the interval $(0, 1)$, and denote the odds $k = (1 - \rho) / \rho$. Theorem \ref{thrm:prob-to-dens} follows as the special case $\rho = 1/2$ with $k=1$. As before, we define $q_0(x) = P_Q(\Delta = 1 |X=x)$, with Bayes' theorem yielding
\begin{align*}
    \alpha_0(x) = \frac{q_0(x)}{1 - q_0(x)} \cdot k,
\end{align*}
and $q_0(x) = \alpha_0(x) / \{k + \alpha_0(x)\}$.

\textbf{Claim}:  Define the transformed generating function
\begin{align*}
    \tilde{F}(t) = \frac{1 - t}{k} \cdot F\left(\frac{t}{1 - t} \cdot k\right).
\end{align*}
Using the shorthand $q(x) \equiv \alpha(x) / \{k+\alpha(x)\}$, we obtain the equality
    \begin{align*}
        \mathcal{R}_F(\alpha)&= \rho^{-1}\E_{Q}\left[ -\tilde{F}\{q(X)\} - \tilde{F}^\prime\{q(X)\}\{\Delta - q(X)\}\right].
    \end{align*}

\begin{proof}
Taking derivatives gives
\begin{align*}
    \tilde{F}^\prime(t) &= \frac{1}{k} F\left(\frac{t}{1-t} \cdot k \right) + \frac{1}{1 - t} F^\prime\left( \frac{t}{1 - t} \cdot k \right)
\end{align*}
Hence we obtain the equalities
\begin{align*}
    \tilde{F}\{q(X)\} &= \frac{F\{\alpha(X)\}}{k + \alpha(X)} \\
    \tilde{F}^\prime\{q(X)\} &= \frac{F\{\alpha(X)\}}{k} + \frac{\{k+\alpha(X)\} F^\prime\{\alpha(X)\}}{k} \\
    q_0(X) - q(X) &= \frac{k\{\alpha_0(X) - \alpha(X)\}}{\{k + \alpha_0(X)\}\{k + \alpha(X)\}}.
\end{align*}
Combining these gives
\begin{align*}
    -\tilde{F}\{q(X)\} - \tilde{F}^\prime\{q(X)\}\{q_0(X) - q(X)\} &= \frac{-F\{\alpha(X)\} - F^\prime\{\alpha(X)\} \{\alpha_0(X) - \alpha(X)\} }{k + \alpha_0(X)}
\end{align*}
The result follows by noting that for a test function $f$
\begin{align*}
    \E_Q[f(X)] &= \E_Q[f(X)|\Delta=0](1 - \rho) + \E_Q[f(X)|\Delta=1] \rho \\
    &= \E_{P_0}[f(X)](1 - \rho) + \E_{P_1}[f(X)]\rho \\
    &= \E_{P_0}[f(X)\{1 - \rho + \rho \alpha_0(X)\}]\\
    &= \rho \E_{P_0}\left[f(X)\left\{k + \alpha_0(X)\right\}\right].
\end{align*}
Hence
\begin{align*}
    \E_{Q}\left[ -\tilde{F}\{q(X)\} - \tilde{F}^\prime\{q(X)\}\{q_0(X) - q(X)\}\right] &= \rho \mathcal{R}_F(\alpha).
\end{align*}
The final equality follows by iterated expectation since $q_0(X) = \E_Q[\Delta\mid X]$.
\end{proof}

\section{Duality of Bregman--Riesz regression and balancing weights}
\label{sect:duality}

Here we present two lemmas, which highlight the connection between balancing weights and Bregman--Riesz regression. In the first, we show that balancing weight learning and Bregman--Riesz regression are Fenchel dual problems. In the second, we give conditions such that the duality gap is zero, and connect the dual solution to the primal solution, assuming that both exist.
We build on related results due to \citet[Proposition E.1]{ben-michael_balancing_2021} and \citet[Lemma 5]{Hirshberg2021}.
% We conclude with an example of $\Lambda$ and $\Lambda^*$ after the proofs below.

\begin{lemma}
\label{lemma1}
Let $\Lambda: \mathcal{H} \to \mathbb{R} \cup \{\infty\}$ be convex with convex conjugate $\Lambda^*$, and with $\Lambda(0) = 0$. Let $F^*$ denote the convex conjugate of $F$. The following two quantities are a primal and dual Fenchel pair
\begin{align}
&\inf_{\alpha \in \mathcal{H}}  \E_{P_0}\left[F\{\alpha(X)\}\right] + \Lambda(\alpha_0 - \alpha), \label{eq:primal} \\
    &\inf_{\beta \in \mathcal{H}}  \E_{P_0}[F^*\{\beta(X)\}] - \h(\beta) + \Lambda^*\left(\beta \right). \label{eq:dual}
\end{align}
\end{lemma}

\begin{lemma}
\label{lemma2}
Assume that: $\E_{P_0}[F\{\alpha_0(X)\}] < \infty$; there exists a minimizer $\tilde{\alpha}$ of the primal problem; and there exists a minimizer $\tilde{\beta}$ of the dual problem.
Then the duality gap is zero, in the sense that \eqref{eq:primal} and \eqref{eq:dual} sum to zero. Also, $(\tilde{\alpha}, \tilde{\beta})$ are unique with $\tilde{\beta}(x) = F^\prime\{\tilde{\alpha}(x)\}$, and
\begin{align}
    \tilde{\alpha} = \argmin_{\alpha \in \mathcal{H}} \mathcal{R}_{F}(\alpha) + \Lambda^*\left(F^\prime \circ \alpha \right). \label{eq:dual_bw}
\end{align}
\end{lemma}

\begin{proof}[Proof of Lemma \ref{lemma1}]
Consider the primal problem
\begin{align*}
    \inf_{\alpha \in \mathcal{H}} s(\alpha) + r(\alpha)
\end{align*}
where
\begin{align*}
    s(\alpha) &\equiv \E_{P_0}\left[F\{\alpha(X)\}\right] \\
    r(\alpha) &\equiv \Lambda(\alpha_0 - \alpha).
\end{align*}
Following \citet[Definition 15.10]{bauschke_convex_2017}, the associated dual problem is
\begin{align*}
    \inf_{\beta \in \mathcal{H}} s^*(\beta) + r^*(-\beta)
\end{align*}
where $s^*$ and $r^*$ are the convex conjugates of $s$ and $r$, which we will derive. Firstly,
\begin{align*}
    s^*(\beta) &= \sup_{\alpha \in \mathcal{H}} \langle \beta, \alpha\rangle - s(\alpha) \\
    &= \sup_{\alpha \in \mathcal{H}} \E_{P_0} \left[ \beta(X) \alpha(X) - F\{\alpha(X)\}\right] \\
    &= \E_{P_0} \left[F^*\{\beta(X)\}\right].
\end{align*}
Secondly,
\begin{align*}
    r^*(\beta) &= \sup_{\alpha \in \mathcal{H}} \langle \beta, \alpha\rangle - \Lambda(\alpha_0 - \alpha) \\
    &= \langle \beta, \alpha_0 \rangle + \sup_{\delta \in \mathcal{H}} \langle -\beta, \delta \rangle - \Lambda(\delta) \\
    &= \langle \beta, \alpha_0 \rangle + \Lambda^*(-\beta)
\end{align*}
where the second line follows by the substitution $\delta = \alpha_0 - \alpha$, and the third from the definition of $\Lambda^*$.
Combining these results, and using the fact that $\langle \beta, \alpha_0 \rangle = \h(\beta)$ we obtain
\begin{align*}
    s^*(\beta) + r^*(-\beta) = \E_{P_0}[F^*\{\beta(X)\}] - \h(\beta) + \Lambda^*\left(\beta \right).
\end{align*}
\end{proof}

\begin{proof}[Proof of Lemma \ref{lemma2}]
Since $F$ is differentiable, $s(\alpha)$ is continuous at $\alpha_0$ with $s(\alpha_0) < \infty$ and $r(\alpha_0) < \infty$. In view of Theorem 15.23 and Proposition 15.24 (vii) of \cite{bauschke_convex_2017}, the duality gap is zero.
Additionally, their Theorem 19.1 implies that $\tilde{\beta}$ is a dual minimizer only if it is an element of the sub differential $\partial s(\tilde{\alpha})$, which we show below is simply the singleton set containing only the function $F^\prime\{\tilde{\alpha}(.)\}$.
Similarly, $\tilde{\alpha}$ is a primal minimizer only if it is an element of the sub differential $\partial s^*(\tilde{\beta})$, which we show is the singleton set containing only $(F^\prime)^{-1}\{\tilde{\beta}(.)\}$. Thus, the primal and dual solution pair is unique.
The expression in \eqref{eq:dual_bw} is obtained by expressing \eqref{eq:dual} in terms of $\beta = F^\prime\{\alpha(.)\}$ and noting that the first two terms are the dual form of the BRR in Remark \ref{remark:duality}.

The statement $\tilde{\beta} \in \partial s(\tilde{\alpha})$ means that for all $\alpha \in \mathcal{H}$
\begin{align*}
    s(\alpha) - s(\tilde{\alpha}) \geq \langle \alpha - \tilde{\alpha}, \tilde{\beta} \rangle.
\end{align*}
Letting $\alpha = \tilde{\alpha} + t \delta$ for $t > 0$ and $\delta \in \mathcal{H}$. This implies that for all $\delta \in \mathcal{H}$
\begin{align*}
    \E_{P_0}\left[ \frac{F\{\tilde{\alpha}(X) + t \delta(X)\} - F\{\tilde{\alpha}(X)\}}{t} \right] \geq \langle \delta, \tilde{\beta} \rangle.
\end{align*}
Since the right does not depend on $t$ the inequality must also be satisfied in the limit $t \to 0$, which exists since $F$ is differentiable. Hence, for all $\delta \in \mathcal{H}$
\begin{align*}
\E_{P_0}\left[ \delta(X) F^\prime\{\tilde{\alpha}(X)\} \right] \geq \langle \delta, \tilde{\beta} \rangle
.
\end{align*}
Hence, this inequality is an equality with $\tilde{\beta}(x) = F^\prime\{\tilde{\alpha}(x)\}$. Following the same argument for the statement $\tilde{\alpha} \in \partial s^*(\tilde{\beta})$ we find that, for all $\delta \in \mathcal{H}$
\begin{align*}
\E_{P_0}\left[ \delta(X) (F^*)^\prime\{\tilde{\beta}(X)\} \right] \geq \langle \delta, \tilde{\alpha} \rangle
.
\end{align*}
Hence, $\tilde{\alpha}(x) = (F^*)^\prime\{\tilde{\beta}(x)\}$, and we note that $(F^*)^\prime$ is $(F^\prime)^{-1}$.
\end{proof}
\end{document}